\documentclass{article}

\PassOptionsToPackage{numbers, compress}{natbib}

\usepackage{amsmath}
\usepackage[preprint]{neurips_2026}

\usepackage{graphicx}
\usepackage[utf8]{inputenc} 
\usepackage[T1]{fontenc}    
\usepackage{hyperref}       
\usepackage{url}            
\usepackage{booktabs}       
\usepackage{amsfonts}       
\usepackage{nicefrac}       
\usepackage{microtype}      
\usepackage{xcolor}         
\usepackage{tabularx}
\usepackage{tcolorbox}
\tcbuselibrary{breakable}
\usepackage{colortbl}
\usepackage{multirow}
\usepackage{comment}
\usepackage{algorithm}
\usepackage{algpseudocode}

\usepackage{caption}
\usepackage{amssymb}
\usepackage{mathtools}
\usepackage{amsthm}
\usepackage[capitalize,noabbrev]{cleveref}
\usepackage{enumitem}
\usepackage{pifont}
\usepackage{xltabular}
\usepackage{marvosym}

\usepackage{subcaption}
\usepackage{array}
\usepackage[table,dvipsnames]{xcolor}
\usepackage{wrapfig}

\makeatletter
\def\blfootnote{\gdef\@thefnmark{}\@footnotetext}
\makeatother

\definecolor{promptgreen}{RGB}{112, 194, 101}
\definecolor{promptbg}{HTML}{F2F9FF}
\definecolor{rowhighlight}{HTML}{ffe6e6}
\definecolor{mydarkred}{RGB}{160, 0, 0}
\definecolor{mutedbluegray}{RGB}{30, 105, 180}
\definecolor{mydarkblue}{RGB}{25, 25, 112}
\definecolor{cvprblue}{rgb}{0.21,0.49,0.74}
\newcommand{\csearch}[1]{\textcolor{NavyBlue}{#1}}
\newcommand{\cinfo}[1]{\textcolor{BrickRed}{#1}}
\newcommand{\ckey}[1]{\textcolor{PineGreen}{#1}}
\newcommand{\cans}[1]{\textcolor{RedViolet}{#1}}
\newcommand{\cthink}[1]{\textcolor{RedOrange}{#1}}

\newcommand{\compactmethodmath}{%
  \setlength{\abovedisplayskip}{3pt plus 0pt minus 1pt}%
  \setlength{\belowdisplayskip}{3pt plus 0pt minus 1pt}%
}

\title{\textit{Don't Guess, Just Ask}: Resolving Ambiguity in Referring Segmentation via Multi-turn Clarification}

\author{
  Yuting Yang\textsuperscript{1,*}\quad 
  Haichao Jiang\textsuperscript{1,*}\quad 
  Tianming Liang\textsuperscript{1}\quad 
  Quan Zhang\textsuperscript{1}\quad 
  Jian-Fang Hu\textsuperscript{1,2,3,\textdagger} \\
  \textsuperscript{1}School of Computer Science and Engineering, Sun Yat-sen University, China \ \ \\
    \textsuperscript{2}Guangdong Province Key Laboratory of Information Security Technology, China \ \ \\
    \textsuperscript{3}Key Laboratory of Machine Intelligence and Advanced Computing, Ministry of Education, China \ \ \\
  \texttt{\{yangyt66, jianghch6, liangtm\}@mail2.sysu.edu.cn} \\
  \texttt{\{zhangq689, hujf5\}@mail.sysu.edu.cn} \\
}

\begin{document}

\maketitle
\blfootnote{\textsuperscript{*}Equal contribution. \quad \textsuperscript{\textdagger}Corresponding author.}

\begin{abstract}
Referring segmentation aims to segment the target objects in images or videos based on the textual query. Despite remarkable progress over the past years, existing works always assume that the user-provided queries are already precise and clear. However, this assumption is impractical. In real-world scenarios, it is unrealistic to expect all users to thoroughly review their visual content and carefully ensure their queries are unique and unambiguous. When encountering such cases, existing segmentation models tend to arbitrarily guess the user preferences, often resulting in undesired outcomes. To address this limitation, we propose \textbf{IC-Seg}, a novel agentic framework that proactively clarifies user intent through multi-turn conversation before segmentation. To effectively incentivize this capability, we further introduce \textbf{Hi-GRPO}, a new hierarchical optimization strategy that injects dense and informative supervision signals at the trajectory, turn, and step levels. This strategy encourages efficient intent clarification, effectively eliminating redundant interactions and improving overall dialogue quality. For evaluation, we establish \textbf{Ambi-RVOS}, a referring video object segmentation benchmark with ambiguous user queries. Extensive experiments demonstrate that IC-Seg not only outperforms existing methods by a large margin in resolving ambiguous queries, but also maintains state-of-the-art performance on standard reasoning segmentation benchmarks. Code and data will be released at \url{https://github.com/iSEE-Laboratory/IC-Seg}.
\end{abstract}

\begin{figure}[t]
  \centering
  \includegraphics[width=1.0\textwidth]{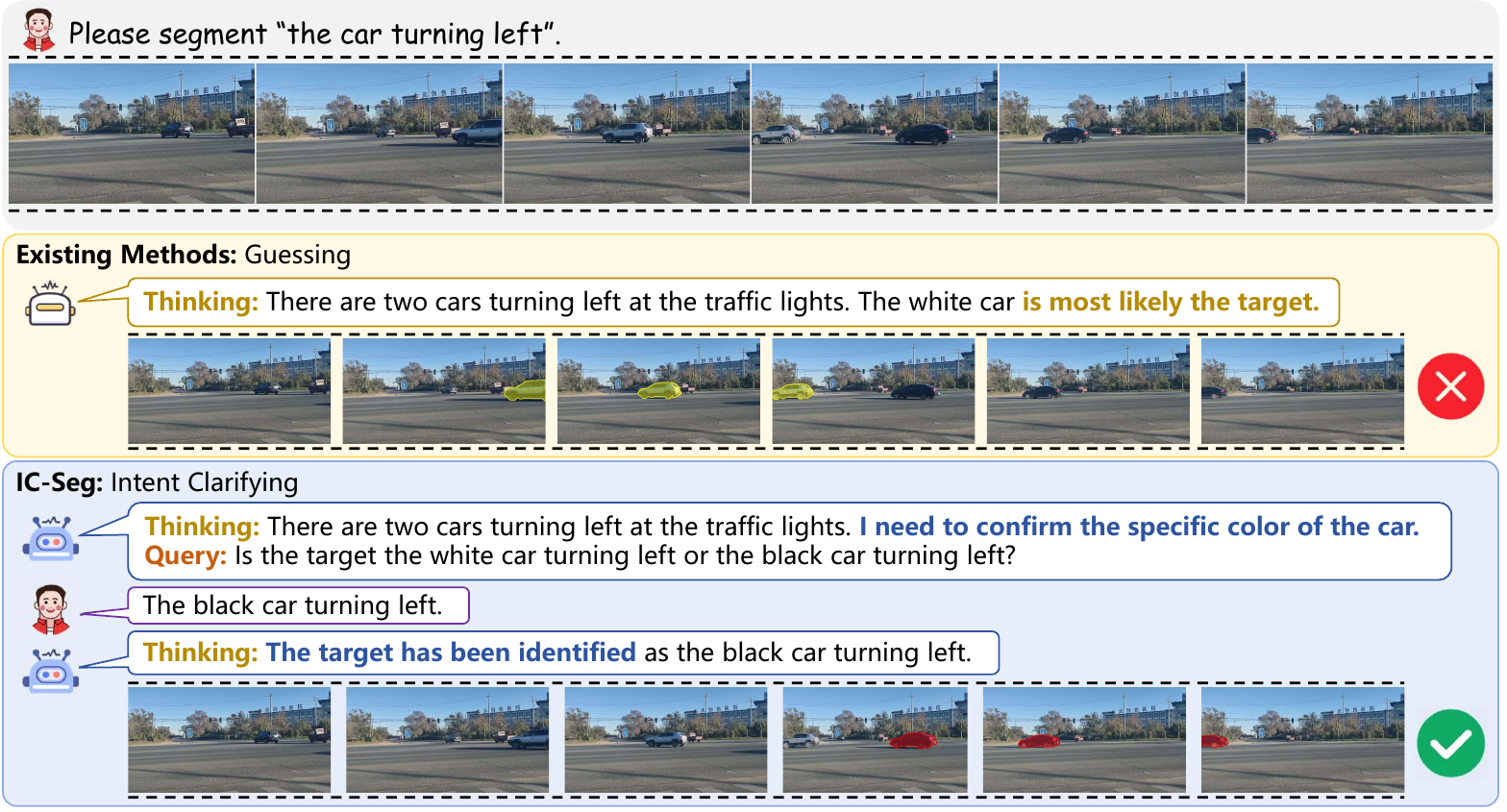}
  \caption{An example of ambiguous referring segmentation. When the user query lacks complete details to uniquely distinguish the intended target, existing methods tend to arbitrarily guess user preferences, while our IC-Seg proactively interacts with the user to clarify their real intention.}
  \label{fig:intro}
  \vspace{-1.5em}
\end{figure}

\section{Introduction}\label{intro}

Recent advancements in Multimodal Large Language Models (MLLMs) have shown great promise in visual perception tasks such as grounding~\cite{Li2025IAGIB, Munasinghe2024VideoGLaMMA} and referring segmentation~\cite{lin2025glus, wang2025object, liu2025unipixel}. Based on the user provided prompts, these models can accurately localize objects of interest in images or videos. This serves as a critical foundation for many downstream applications, including video editing, autonomous driving, and human-robot interaction.


Despite these impressive advancements, existing approaches~\cite{liang2025referdino, liu2025seg, gong2026veason} typically assume that user queries are precise and clear, containing all necessary details to uniquely distinguish the expected target. However, this assumption is highly impractical, as real-world user instructions are often ambiguous and imperfect. For example, a user might simply ask to "segment the red car" without noticing that there is also another car with a similar color in the video. The demand for addressing such ambiguous queries is widespread, as it is unrealistic to expect all users to exhaustively check the visual content and carefully ensure their referring expressions are unambiguous.


To bridge this gap, we propose \textbf{IC-Seg}, a trainable agentic framework for Referring \textbf{Seg}mentation with multi-turn \textbf{I}ntent \textbf{C}larification. As shown in Figure \ref{fig:intro}, when encountering ambiguous queries, unlike existing models that perform unreliable guessing, IC-Seg proactively interacts with the users to confirm their intent and preferences, thereby ensuring accurate prediction for the desired target.


However, training such a multi-turn, long-horizon interactive capability is still challenging. While the recent reinforcement learning algorithms~\cite{rafailov2023direct, yu2025dapo} like Group Relative Policy Optimization (GRPO)~\cite{shao2024deepseekmath} provide potential in LLM post training, our preliminary study empirically reveals that these approaches would lead the model towards generating repetitive and uninformative conversation, rather than actively resolving the ambiguity (see Figure~\ref{fig:visual}). To address this limitation, we introduce \textbf{Hi-GRPO}, a \textbf{Hi}erarchical optimization strategy that injects structured supervision signals at three different levels: (i) at the \textit{trajectory} level, it rewards the responses that achieve higher final accuracy; (ii) at the \textit{turn} level, it encourages the intermediate actions that effectively shrinks the query ambiguity; (iii) at the \textit{step} level, it reinforces the specific tokens that contribute to the success of the long-horizon decision-making process. This hierarchical training strategy provides dense and informative supervision signals, effectively incentivizing the multi-turn proactive interaction abilities of IC-Seg.


To assess the effectiveness of IC-Seg, we further establish Ambi-RVOS, a new benchmark for Referring Video Object Segmentation that specifically necessitates Intent Clarification. Unlike existing benchmarks that assume all user queries are precise and unique, Ambi-RVOS is designed to reflect the inherent ambiguity of human instructions in real-world scenarios. Experimental results reveal that existing methods struggle severely with ambiguous user queries, while our IC-Seg demonstrates superior performance through strategic interactions, yielding nearly 20\% improvements in overall $\mathcal{J}\&\mathcal{F}$. Furthermore, IC-Seg preserves state-of-the-art (SOTA) performance on existing image and video reasoning segmentation benchmarks, demonstrating its overall robustness.


Our main contributions are summarized as follows:

\begin{itemize}[leftmargin=15pt, labelindent=0pt, itemindent=0pt]
\item We propose IC-Seg, an agentic language-guided segmentation framework that performs multi-turn interactive clarification to resolve ambiguity prior to segmentation.
\item We design Hi-GRPO, a Hierarchical Group Relative Policy Optimization strategy that utilizes a three-level supervisory signals to foster effective multi-turn clarification capabilities.
\item We establish Ambi-RVOS, a benchmark specifically designed to reflect the inherent ambiguity of human instructions in real-world scenarios.
\item Extensive experiments demonstrate that IC-Seg achieves SOTA performance on both Ambi-RVOS benchmark and existing image/video reasoning segmentation benchmarks.
\end{itemize}

\section{Related Work}
\textbf{Language-guided Segmentation.}
Referring Segmentation~\cite{ding2023mevis, seo2020urvos} aims to identify and segment target objects in images or videos based on natural language descriptions. While early approaches~\cite{botach2022end, luo2023soc, miao2023spectrum, liang2025longrvos} relied on deterministic matching frameworks (e.g., DETR~\cite{carion2020end}) for straightforward queries, recent advancements~\cite{lin2025glus, liu2025unipixel, jiang2026refer} have increasingly integrated Multi-modal Large Language Models (MLLMs) to handle complex reasoning. Pioneering works~\cite{lai2024lisa, yan2024visa, bai2024one} introduced Reasoning Segmentation, proposing a special \texttt{<SEG>} token to bridge MLLMs and foundation segmentation models for implicit instructions. To further unleash the reasoning potential of MLLMs in complex scenes, recent efforts~\cite{xu2026videoseg, gong2026veason} have employed reinforcement learning (RL) to optimize policy MLLMs for generating spatial prompts. However, these paradigms remain fundamentally \textit{passive} as they rely on a single-pass inference pipeline, lacking the capability to proactively interact with users or the environment to seek necessary confirmation. To bridge this gap toward practical human-centric applications, we introduce IC-Seg, a framework that empowers models with intent clarification capabilities to resolve reference ambiguities, alongside Ambi-RVOS, a novel benchmark demanding interaction and confirmation for accurate segmentation.


\textbf{Multi-turn Visual Interaction.}
Human-in-the-loop visual perception has long been studied as a way to incorporate user feedback into model predictions. Visual dialog and visual question answering~\cite{das2017visual, de2017guesswhat, wu2021fashion} enable multi-turn language interaction about images or videos, requiring models to understand visual content, track dialogue history, and generate context-aware responses. Recent MLLMs further extend this capability to open-ended visual conversations~\cite{liu2023visual, chen2023llava, bai2025qwen3vl, ni2026survey}, making natural-language interaction a practical interface for visual understanding. However, most existing multi-turn visual interaction remains response-driven: the model answers the user's current request, but does not explicitly verify whether the underlying intent has been specified clearly enough. IC-Seg instead treats dialogue as a process of intent clarification, where the model proactively identifies ambiguity, asks targeted questions, and uses the clarified intent for final grounding.

\textbf{Reinforcement Learning with LLMs.}
Reinforcement Learning with Verifiable Rewards (RLVR) has become an effective paradigm for improving the reasoning ability of LLMs by optimizing them with checkable outcome rewards. Recent algorithms such as Group Relative Policy Optimization (GRPO)~\cite{shao2024deepseekmath} further improve training efficiency by estimating relative advantages within sampled groups without relying on a learned value model. Beyond single-turn reasoning, Agentic RL extends RL training to interactive settings where models make multi-step decisions and interact with environment~\cite{jin2025searchr, zheng2025stepsearch, geng2025webwatcher, liang2026seg}. This broader setting introduces a more challenging credit assignment problem, as a sparse outcome reward often provides limited information about intermediate actions.
This motivates training signals that are more informative than a single terminal reward. On-Policy Distillation (OPD)~\cite{agarwal2024policy, yang2025qwen3, li2026rethinking, hubotter2026reinforcement, yang2026self} provide one relevant perspective by re-evaluating sampled outputs under teacher or privileged conditions to obtain token-level guidance. However, these token-level signals can be noisy if used as standalone optimization targets.
These observations motivate our Hi-GRPO design, which combines task-specific rewards with token-level supervision for agentic interaction.

\begin{figure}
  \centering
  \includegraphics[width=.9\textwidth]{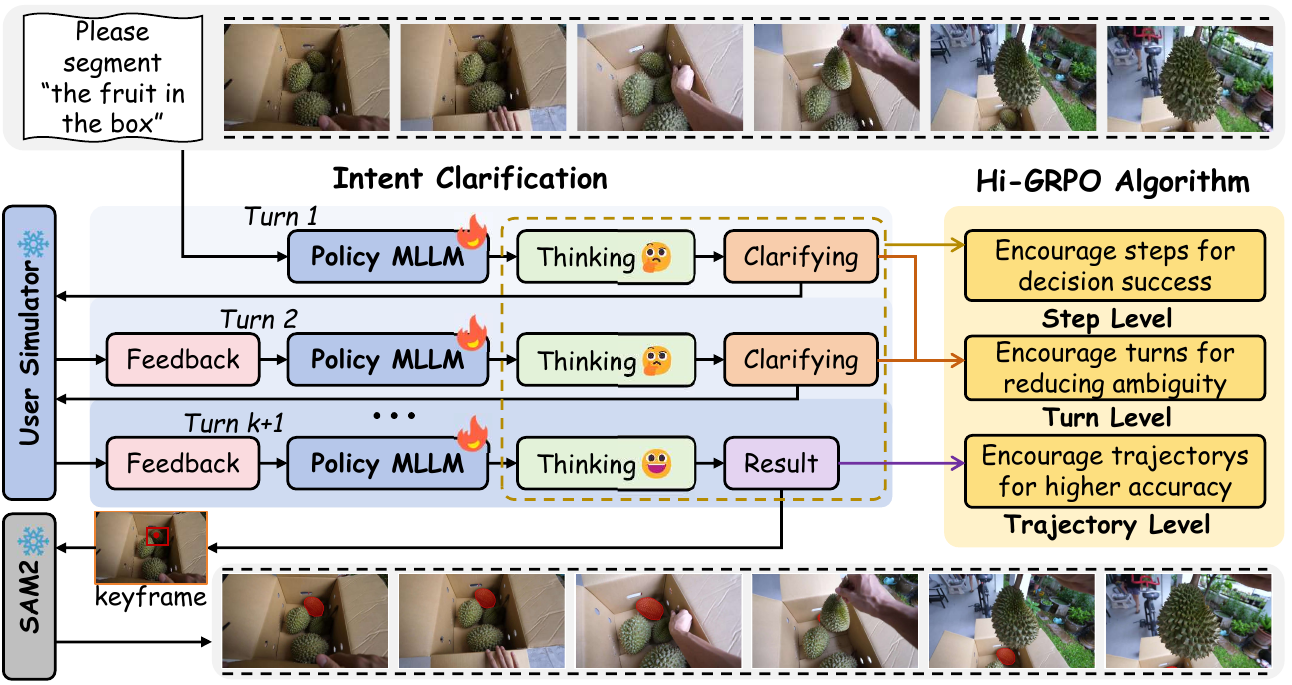}
  \caption{Overview of the IC-Seg framework. IC-Seg resolves ambiguities via multi-turn dialogues with an MLLM-based User Simulator. Our Hi-GRPO algorithm empowers the agent through a hierarchical reward chain: supervising final localization accuracy at the trajectory level, inquiry quality at the turn level, and fine-grained reasoning steps using expert-diagnosed signals.}
  \label{fig:method}
  \vspace{-1.5em}
\end{figure}

\begingroup
\compactmethodmath
\section{IC-Seg: Referring Segmentation with Multi-turn Intent Clarification}
IC-Seg is an agentic segmentation framework that aims to segment the object targets from images or videos based on the textual query. The most prominent feature is, when the query is ambiguous, IC-Seg is able to leverage multi-turn dialogue to clarify the user's real intent, rather than making unreliable guesses, as illustrated in \Cref{fig:method}. This complex capability is effectively incentivized by our Hierarchical Group Relative Policy Optimization (Hi-GRPO), which provides dense and informative supervision signals to each action trajectory at three different levels. In the following sections, we mainly elaborate on the pipeline of IC-Seg and the design of Hi-GRPO.

\subsection{Clarification-then-Segmentation Pipeline}\label{pipeline}
Unlike existing RVOS paradigms that assume user queries are always precise and unique, IC-Seg is designed to proactively identify and resolve query ambiguity through multi-turn dialogue with the user. As illustrated in \Cref{fig:method}, our framework formulates this process into a dynamic, multi-turn interaction loop driven by a trainable policy MLLM $\pi_\theta$. Once the user intent has been confirmed, $\pi_\theta$ will proceed to localize the target and perform segmentation.


\textbf{Clarification loop.} Given a visual input (an image or video) and a textual query, $\pi_\theta$ first reasons over the visual-textual inputs. During each iteration, if $\pi_\theta$ believes that the user intent remains ambiguous, it generates a \textit{clarifying question} to ask for essential details from the user. In order to simulate the human-in-the-loop process during online training, we introduce an additional MLLM $\pi_B$ as the User Simulator. This simulator, provided with the ground-truth target and the clarifying question, generates \textit{user feedback} in response to $\pi_\theta$. This clarification-feedback dialogue can iterate from zero to multiple times until the ambiguity is fully resolved. 


\textbf{Segmentation.} Once the user intent has been confirmed, $\pi_\theta$ proceeds to localize the target objects following existing paradigms~\cite{xu2026videoseg, gong2026veason, liang2026seg}. Specifically, for video tasks, $\pi_\theta$ first selects a keyframe on which the target is most prominently visible, and outputs a bounding box together with a point inside the target on that keyframe as the positional prompt. This prompt is finally fed to a frozen SAM2 for mask generation over the whole video. For image tasks, the keyframe selection is naturally omitted.

\subsection{Hierarchical Group Relative Policy Optimization for Multi-turn Interaction}
To ensure the policy MLLM $\pi_\theta$ effectively resolves ambiguity through dialogue, we must supervise not only the final localization accuracy but also the strategic quality of the interactive process. To this end, we propose Hi-GRPO that provides dense and effective supervision signals at the trajectory, turn, and step levels. Following the core principle of GRPO, Hi-GRPO maximizes the following objective:
\begin{equation}
\label{eq:grpo}
\mathcal{L} (\theta )=\mathbb{E} \left[ \frac{1}{G}\sum_{i=1}^G{\frac{1}{|y_i|}}\sum_{t=1}^{|y_i|}{\min \left( \rho_{i,t}\tilde{A}_{i,t}, \text{clip}\left( \rho_{i,t},1-\epsilon ,1+\epsilon \right) \tilde{A}_{i,t} \right)} \right],
\end{equation}
where $\{y_1, y_2, \dots, y_G\}$ denotes a group of $G$ trajectories sampled from the old policy $\pi_{\theta_{\text{old}}}$. Here, $\rho _{i,t}=\frac{\pi _{\theta}(y_{i,t}|x,y_{i,<t})}{\pi _{\theta _{\text{old}}}(y_{i,t}|x,y_{i,<t})}$ is the importance sampling ratio, $x$ represents the input, $y_{i,t}$ denotes the $t$-th token in the $i$-th sampled trajectory, and $\epsilon$ is the clipping threshold.
In the GRPO framework, the advantage $\tilde{A}$ is derived solely from the final outcome of the model and is uniformly assigned to each token throughout the trajectory, which provides sparse supervision that causes the model to overlook interaction quality.To address this, Hi-GRPO deconstructs the advantage into a hierarchical structure, capturing optimization signals across three distinct levels. We next detail the design of these levels and their integration into the final advantage signal. The pseudocode is presented in Appendix~\ref{sec:code}.

\subsubsection{Trajectory-Level Optimization.}
Trajectory-level optimization aims to maximize the accuracy of model's task completion. Following common practice in MLLM-based VOS~\cite{xu2026videoseg, gong2026veason, liang2026seg}, we supervise the model's keyframe-level localization output by optimizing the policy against four complementary terms:
\begin{equation}\label{eq:rtraj}
    \mathcal{R}^{\text{traj}} = \mathcal{R}^{\text{iou}} + \mathcal{R}^{\text{box}} + \mathcal{R}^{\text{point}} + \mathcal{R}^{\text{keyframe}},
\end{equation}
where $\mathcal{R}^{\text{iou}}$ and $\mathcal{R}^{\text{box}}$ are binary optimization constraints that reward an $\text{IoU} > 0.5$ and a center $L_1$ distance $< 10$ pixels, respectively. The $\mathcal{R}^{\text{point}}$ incentivizes accurate point prediction by yielding a positive signal if the output falls within both the bounding box and a $100$-pixel radius of the ground truth. Finally, the continuous term $\mathcal{R}^{\text{keyframe}}$ encourages the selection of visually clear frames, calculated as the ratio of the target's current mask area to its maximum area across the video.


\subsubsection{Turn-Level Optimization.}
With only the outcome reward, the model often terminates prematurely despite unresolved ambiguities, or falls into uninformative redundancy by issuing repetitive queries that yield no information gain. We thus introduce turn-level optimization to improve dialogue quality from two aspects: ambiguity reduction and interaction efficiency. The reward is computed as follows:
\begin{equation}
  \mathcal{R}^{\text{turn}} = \mathcal{R}^{\text{ent}} + \mathcal{R}^{\text{eff}},
\end{equation}
\textbf{Entropy Reduction.} 
To prevent premature termination of the dialogue, we quantify the ambiguity of the referring expression using Shannon entropy and incentivize the maximization of information gain. Let $N_{k}$ denote the size of the candidate set after the $k$-th interaction turn, where $N_{0} = M$ represents the initial number of candidates and $N_{K} = N$ represents the residual count following the final turn $K$. Under a uniform prior over the candidate set, the turn-$k$ entropy is $H_{k} = -\sum_{i=1}^{N_{k}} p_{i}\log_{2}p_{i} = \log_{2}N_{k}$ with $p_i=1/N_k$ and $N_k>0$. The entropy reduction objective, $\mathcal{R}^{\text{ent}}$, is then defined by normalizing the total entropy reduction as follows:
\begin{equation}
  \mathcal{R}^{\text{ent}} = \frac{H_0 - H_K}{H_0}, \quad H_0 > 0.
\end{equation}
Since the candidate-set size $N_k$ is not directly observable during the rollout, we employ the model $\pi_B$ as an automated judge upon the completion of each episode. To facilitate this evaluation, the judge is provided with the video frames, textual descriptions of all initial candidate objects, and the full clarification dialogue between $\pi_\theta$ and $\pi_B$. The judge is then instructed to simulate the interaction turn-by-turn to estimate the progressive elimination of candidates, ultimately outputting the sequence of residual candidate counts $\mathbf{c} = [N_1, N_2, \dots, N_K]$.

\textbf{Interaction Efficiency.} To penalize redundant or ineffective inquiries that fail to prune the candidate set, we define the efficiency optimization objective $\mathcal{R}^{\text{eff}}$. Let $K$ denote the total number of clarification questions asked in a rollout. We use the per-turn candidate counts $\mathbf{c} = [N_1, N_2, \dots, N_K]$ and declare the $k$-th question effective whenever $N_{k} < N_{k-1}$. The objective is defined as:
\begin{equation}\label{eq:reff}
\mathcal{R}^{\text{eff}} = \frac{1}{K} \sum_{k=1}^{K} \mathbb{I}(N_{k} < N_{k-1}),
\end{equation}
where $\mathbb{I}(\cdot)$ is the indicator function. By normalizing the count of informative questions by the total number of turns $K$, this term encourages the model to ask concise and discriminative questions.

\subsubsection{Step-Level Optimization}
While trajectory and turn optimization provide supervision for the model's localization and interaction processes, their rewards are still sparse, lacking the ability to identify which specific steps contribute to the success or failure of long-horizon decision-making. Inspired by recent OPSD works~\cite{hubotter2026reinforcement, yang2026self}, we introduce step-level optimization to provide dense supervision. For each sampled trajectory, we construct two views of $\pi_{\theta}$. 1) The \textit{student view} conditions only on the original prompt and measures how likely the policy is to generate each sampled token. 2) The \textit{self-teacher view} conditions on the same input plus additional privileged guidance $r$, and re-evaluates the likelihood of the identical token sequence with $r$. Comparing these two likelihoods reveals how much each token benefits from the guidance, thereby providing fine-grained signals for step-level credit assignment.

Existing OPSD methods commonly use ground-truth answers or traces as privileged guidance. This is insufficient for IC-Seg, where success depends not only on the final target, but also on whether the model asks informative questions and terminates the dialogue at the appropriate time.
To bridge this gap, we utilize the model $\pi_{\text{B}}$—the same automated judge used for entropy estimation—to provide expert diagnostic feedback. For each sampled trajectory, the $\pi_{\text{B}}$ analyzes the model's quality in interaction and localization. For example, if the model's inquiry is ambiguous, $\pi_{\text{B}}$ explicitly identifies the source of confusion and incorporates these cautions into $r$.
By conditioning on this guidance $r$, the self-teacher re-evaluates the likelihood of its previously generated tokens. The ratio between the privileged and the original probabilities is defined as:
\begin{equation}
f_t=\frac{\pi _{\theta}\left( y_t|x,r,y_{<t} \right)}{\pi _{\theta}\left( y_t|x,y_{<t} \right)},
\end{equation}
where $x$ represents the input video and initial prompt. This ratio $f_t$ serves as a fine-grained indicator of token quality. By incorporating $f_t$ into the final optimization $\mathcal{L}(\theta)$ as defined in Eq. \ref{eq:grpo}, we force the original policy's probability distribution to approach the privileged distribution during training.

\subsubsection{Hierarchical Advantage Integration}
Integrating the objectives from the three levels, we formulate our final hierarchical advantage function:
\begin{equation}
    \tilde{A}_{i,t} = \underbrace{\frac{\mathcal{R} _{i}^{\mathrm{traj}}+\alpha \cdot \mathcal{R} _{i}^{\mathrm{turn}}-\mu _G}{\sigma _G}}_{\text{Trajectory+Turn Optimization}} \cdot \underbrace{\left( (1-\lambda )+\lambda \cdot f_{i,t}^{s_i} \right)}_{\text{Step Optimization}} ,
\end{equation}
where $\alpha$ is a coefficient of the turn-level objective. The $\mu_G$ and $\sigma_G$ represent the mean and standard deviation of the combined objectives $(\mathcal{R}_{i}^{\text{traj} } + \alpha \cdot \mathcal{R}_{i}^{ \text{turn} })$ across the sampled group of $G$ trajectories. The $\lambda \in [0, 1]$ controls the influence intensity of the step-level optimization. The $s_i$ is a directional indicator, where $s_i = 1$ if the Trajectory+Turn Optimization term is positive and $s_i = -1$ otherwise.

 The trajectory and turn optimization serves as a sequence-level operator, evaluating the $i$-th trajectory against the group $G$ to dictate the global update direction. Conversely, the step-level optimization term modulates the update magnitude for each token $t$ based on its alignment with privileged expert, without altering the overall direction. Compared to the outcome-based advantage in vanilla GRPO, this hierarchical advantage effectively supervises and guides the model's clarification process.

\endgroup

\section{Ambi-RVOS: An RVOS Benchmark with Ambiguous User Queries}
To evaluate the intent clarification capabilities of models, we introduce Ambi-RVOS, a benchmark for Referring Video Object Segmentation that contains Ambiguous user queries. Unlike conventional RVOS benchmarks~\cite{bai2024one, seo2020urvos} that assume the user queries are precise and unique, Ambi-RVOS presents a fundamentally different challenge where a query maps to multiple candidates, but only one specific individual is the intended ground truth.
To construct this benchmark, we curated videos from the ReVOS~\cite{yan2024visa} and MeViS~\cite{ding2023mevis} datasets, focusing on scenes with a high density of visually similar objects. We obtain ambiguous queries through two procedures. First, when the source query already matches multiple objects in the scene, we directly preserve it. Second, when the query uniquely identifies one target with detailed attributes, we progressively remove attributes and details until the simplified query refers to more than one candidate.

The Ambi-RVOS benchmark consists of 1500 curated test samples, each comprising a video sequence and an ambiguous query.
We partition the test set into three complexity levels based on the initial number of candidate objects associated with the ambiguous query: Simple, Medium and Difficult.
Specifically, the Simple category comprises 400 samples featuring exactly two candidate objects; the Medium category includes 600 samples with three to five objects; and the Difficult category contains 500 samples involving complex scenes with six or more objects. This distribution ensures a comprehensive assessment across varying levels of reference ambiguity. By transitioning from traditional passive visual recognition to interactive reasoning, Ambi-RVOS facilitates a paradigm shift in RVOS toward more human-centric and practical applications.

\begin{table*}[t]
\setlength\tabcolsep{5.5pt}
\footnotesize
\centering
\caption{Comparison on Ambi-RVOS benchmark.}
\vspace{-0.5em}
\label{tab:Ambi-RVOS}
\begin{tabular}{l | ccc| ccc| ccc| ccc}
\toprule
\multirow{2}{*}{Method} & \multicolumn{3}{c|}{\textbf{Simple}} & \multicolumn{3}{c|}{\textbf{Medium}} & \multicolumn{3}{c|}{\textbf{Difficult}} & \multicolumn{3}{c}{\textbf{Overall}} \\
\cmidrule(lr){2-13}
 & $\mathcal{J}\&\mathcal{F}$ & $\mathcal{J}$ & $\mathcal{F}$ & $\mathcal{J}\&\mathcal{F}$ & $\mathcal{J}$ & $\mathcal{F}$ & $\mathcal{J}\&\mathcal{F}$ & $\mathcal{J}$ & $\mathcal{F}$ & $\mathcal{J}\&\mathcal{F}$ & $\mathcal{J}$ & $\mathcal{F}$ \\
\midrule

\multicolumn{13}{l}{\textit{\textbf{Specialist RVOS models}}} \\
SAMWISE~\cite{cuttano2025samwise} & 29.8 & 26.3 & 33.4 & 19.0 & 16.5 & 21.5 & 11.7 & 9.4 & 14.0 & 19.4 & 16.7 & 22.2 \\
ReferDINO~\cite{liang2025referdino} & 33.2 & 29.9 & 36.5 & 20.0 & 17.7 & 22.2 & 13.4 & 11.5 & 15.3 & 21.3 & 18.9 & 23.7 \\
SAM3~\cite{carion2025sam3} & 21.6 & 20.9 & 22.3 & 19.0 & 17.0 & 21.0 & 15.8 & 14.7 & 17.0 & 18.6 & 17.3 & 20.0 \\
\midrule

\multicolumn{13}{l}{\textit{\textbf{MLLM-based methods}}} \\
GLUS-7B~\cite{lin2025glus} & 37.8 & 33.9 & 41.8 & 27.3 & 22.7 & 31.8 & 18.1 & 14.7 & 21.6 & 27.1 & 23.0 & 31.1 \\
RGA3-7B~\cite{wang2025object} & 37.3 & 33.0 & 41.5 & 26.3 & 21.9 & 30.8 & 12.8 & 9.4 & 16.3 & 24.7 & 20.7 & 28.8 \\
UniPixel-7B~\cite{liu2025unipixel} & 42.4 & 37.9 & 46.8 & 27.7 & 23.3 & 32.2 & 20.3 & 16.6 & 24.0 & 29.2 & 25.0 & 33.3 \\
Veason-R1-3B~\cite{gong2026veason} & 22.6 & 19.1 & 26.2 & 15.1 & 12.2 & 18.0 & 16.0 & 14.0 & 17.9 & 17.4 & 14.7 & 20.2 \\
\midrule

Qwen3-VL-4B* & 35.8 & 33.5 & 38.0 & 31.6 & 29.8 & 33.5 & 24.5 & 23.1 & 26.0 & 30.4 & 28.6 & 32.2 \\
\rowcolor{rowhighlight}
\textbf{IC-Seg-4B (ours)} & \textbf{63.3} & \textbf{60.3} & \textbf{66.2} & \textbf{54.3} & \textbf{51.0} & \textbf{57.6} & \textbf{40.3} & \textbf{38.0} & \textbf{42.6} & \textbf{52.0} & \textbf{49.2} & \textbf{54.9} \\
\midrule

Qwen3-VL-8B* & 43.8 & 40.6 & 47.0 & 38.6 & 36.1 & 41.1 & 28.2 & 25.8 & 30.6 & 36.5 & 33.9 & 39.2 \\
\rowcolor{rowhighlight}
\textbf{IC-Seg-8B (ours)} & \textbf{63.7} & \textbf{60.3} & \textbf{67.0} & \textbf{57.4} & \textbf{54.1} & \textbf{60.8} & \textbf{45.5} & \textbf{42.6} & \textbf{48.3} & \textbf{55.1} & \textbf{51.9} & \textbf{58.3} \\
\bottomrule
\end{tabular}\vspace{-1em}
\end{table*}

\begin{table*}[t]
  \centering
  \begin{minipage}[t]{0.48\textwidth}
    \centering
    \caption{Comparison on ReasonSeg benchmark.}
    \vspace{-0.5em}
    \label{tab:reasonseg}
    \setlength{\tabcolsep}{6pt}
    \footnotesize
    \begin{tabular}{l|cc|cc}
      \toprule
      \multirow{2}{*}{Method} & \multicolumn{2}{c|}{Val} & \multicolumn{2}{c}{Test} \\
       \cmidrule(lr){2-3} \cmidrule(lr){4-5}
       & gIoU & cIoU & gIoU & cIoU \\
      \midrule
      LISA-7B~\cite{lai2024lisa} & 53.6 & 52.3 & 48.8 & 47.1 \\
      Seg-Zero-7B~\cite{liu2025seg} & 62.6 & 62.0 & 57.5 & 52.0 \\
      SAM-R1-7B~\cite{huang2025samr} & 64.0 & 55.8 & 60.2 & 54.3 \\
      SAM3-Agent-7B~\cite{carion2025sam3} & 65.4 & 50.5 & 62.6 & 56.2 \\
      \midrule
      Qwen3-VL-8B* & 70.6 & 71.2 & 65.4 & 54.7 \\
      \rowcolor{rowhighlight}
      \textbf{IC-Seg-8B (ours)} & \textbf{72.3} & \textbf{74.6} & \textbf{66.7} & \textbf{60.2} \\
      \bottomrule
    \end{tabular}
  \end{minipage}%
  \hfill
  \begin{minipage}[t]{0.49\textwidth}
    \centering
    \caption{Comparison on ReasonVOS benchmark.}
    \vspace{-0.5em}
    \label{tab:reasonvos}
    \setlength{\tabcolsep}{10pt}
    \footnotesize
    \begin{tabular}{l | ccc}
      \toprule
      Method & $\mathcal{J}\&\mathcal{F}$ & $\mathcal{J}$ & $\mathcal{F}$  \\ [4.8pt]
      \midrule
      VideoLISA-3.8B~\cite{bai2024one} & 47.5 & 45.1 & 49.9 \\
      GLUS-7B~\cite{lin2025glus} & 49.9 & 47.5 & 52.4 \\
      RGA3-7B~\cite{wang2025object} & 53.6 & 51.3 & 56.0 \\
      COT-RVS-12B~\cite{kao2025cot} & 50.7 & 47.5 & 54.0 \\
      OneThinker-8B~\cite{feng2025onethinker} & 54.9 & 51.1 & 58.7 \\
      \midrule
      Qwen3-VL-8B* & 56.2 & 53.1 & 59.2 \\
      \rowcolor{rowhighlight}
      \textbf{IC-Seg-8B (ours)} & \textbf{60.9} & \textbf{57.5} & \textbf{64.2} \\
      \bottomrule
    \end{tabular}
  \end{minipage}
  \vspace{-1em}
\end{table*}

\section{Experiments} \label{sec:exp}
\subsection{Experimental Setup}\label{sec:exp setup}
\textbf{Implementation Details.}
We adopt Qwen3-VL-Instruct-4B and 8B as our base MLLMs, and use SAM2-Large~\cite{ravi2024sam} to produce the per-frame precise object masks. We manually collect and annotate 120 samples for RL training. We use Qwen3-VL-Instruct-32B as the user simulator. $\alpha$ is set to 0.5 and $\epsilon$ is set to 0.2. Following RLSD~\cite{yang2026self}, $\lambda$ is initialized at 0.5 but linearly decayed to 0 during the whole training process, and $\epsilon_f$ is set to 0.2.
More details are present in the Appendix~\ref{sec:more_details}. 

\textbf{Evaluation Metrics.}
Our metrics follow the previous works~\cite{ding2023mevis, liu2025seg}. For video object segmentation, we use region similarity $\cal{J}$ (average IoU), contour accuracy $\cal{F}$ (mean boundary similarity), and average $\cal{J\&F}$. For image segmentation, we use gIoU (generalized IoU) and cIoU (cumulative IoU).

\textbf{Baselines.}
 For automatic evaluation on Ambi-RVOS, we apply a Qwen3-VL-32B as the user simulator to provide feedback. Each method is given the same ambiguous query, visual input, and access to the user simulator; whether and how to initiate clarification is determined by the method itself. We implement Qwen3-VL* as the baseline by pairing the corresponding Qwen3-VL-Instruct backbone with the same mask generator as IC-Seg, but without our Hi-GRPO training. 

\subsection{Main Results}\label{sec:main results}
\textbf{Comparison on Ambi-RVOS.} Table~\ref{tab:Ambi-RVOS} presents the quantitative evaluations on our Ambi-RVOS benchmark. The results reveal that traditional specialist RVOS models (e.g., ReferDINO~\cite{liang2025referdino}, $21.3$ $\mathcal{J}\&\mathcal{F}$) and advanced reasoning MLLMs (e.g., UniPixel-7B~\cite{liu2025unipixel}, $29.2$ $\mathcal{J}\&\mathcal{F}$) all struggle on this task, as they lack the proactive clarifying mechanism to resolve reference ambiguities. The Qwen3-VL* baseline improves over non-interactive models but still struggles to fully exploit user feedback. By contrast, our IC-Seg effectively addresses this gap and achieves the best performances across all complexity levels. Notably, IC-Seg-4B delivers a $+21.6$ $\mathcal{J}\&\mathcal{F}$ improvement over Qwen3-VL-4B* baseline, while IC-Seg-8B establishes a new state-of-the-art performance of $55.1$ $\mathcal{J}\&\mathcal{F}$, proving that our method successfully enables the model to resolve ambiguity through intent interaction.

\textbf{Comparison on Reasoning Benchmarks.} Beyond interactive scenarios, we verify our model's fundamental visual grounding proficiency on the ReasonSeg and ReasonVOS datasets (Table~\ref{tab:reasonseg} and Table~\ref{tab:reasonvos}). Despite being trained for multi-turn clarification, IC-Seg achieves leading metrics on both benchmarks. Notably, IC-Seg-8B substantially outperforms recent competitors like SAM3-Agent-7B~\cite{carion2025sam3} ($+4.1$ Test gIoU) and OneThinker-8B~\cite{feng2025onethinker} ($+6.0$ $\mathcal{J}\&\mathcal{F}$), even though OneThinker-8B is built upon the identical Qwen3-VL-8B base model. These results confirm that our training framework preserves and even strengthens the core spatial-temporal reasoning capabilities of the policy MLLM.

\begin{table}[t]
  \centering
  \begin{minipage}[t]{0.47\textwidth}
    \centering
    \caption{Effects of the three-level design in Hi-GRPO: $\mathcal{R}_{ent}$, $\mathcal{R}_{eff}$ and $\mathcal{R}_{step}$.}
    \label{tab:hi-grpo}
    \setlength{\tabcolsep}{6pt}
    \footnotesize
    \begin{tabular}{ccc|ccc}
      \toprule
       $\mathcal{R}_{ent}$ & $\mathcal{R}_{eff}$ & $\mathcal{R}_{step}$ & $\mathcal{J}\&\mathcal{F}$ & $\mathcal{J}$ & $\mathcal{F}$ \\
      \midrule
      \rowcolor{gray!7.5}
       & & & 46.4 & 43.3 & 49.4 \\
      \ding{51} & & & 47.5 & 44.3 & 50.6 \\
       & \ding{51} & & 48.3 & 45.2 & 51.5 \\
       & & \ding{51} & 48.5 & 45.7 & 51.2 \\
      \ding{51} & \ding{51} & & 48.9 & 45.9 & 52.0 \\
      \rowcolor{rowhighlight}
      \ding{51} & \ding{51} & \ding{51} & \textbf{52.0} & \textbf{49.2} & \textbf{54.9} \\
      \bottomrule
    \end{tabular}
  \end{minipage}%
  \hfill
  \begin{minipage}[t]{0.49\textwidth}
    \centering
    \caption{Impact of the privileged context $r$. ``Guidance'' indicates our expert guidance. ``Empty'' denotes no context. ``Answer'' uses the ground-truth.}
    \label{tab:feedback}
    \setlength{\tabcolsep}{8pt}
    \footnotesize
    \begin{tabular}{c|c|ccc}
      \toprule
       Method & r & $\mathcal{J}\&\mathcal{F}$ & $\mathcal{J}$ & $\mathcal{F}$ \\ [2.5pt]
      \midrule
      \rowcolor{gray!7.5}
      SDPO~\cite{hubotter2026reinforcement} & Guidance & 21.3 & 19.3 & 23.2 \\
      \midrule
      Hi-GRPO & Empty & 48.7 & 46.0 & 51.4 \\
      Hi-GRPO & Answer & 49.9 & 46.9 & 52.9 \\
      \rowcolor{rowhighlight}
      Hi-GRPO & Guidance & \textbf{52.0} & \textbf{49.2} & \textbf{54.9} \\
      \bottomrule
    \end{tabular}
  \end{minipage}
  \vspace{-1em}
\end{table}

\begin{table}[t]
  \centering
  \begin{minipage}[t]{0.48\textwidth}
    \centering
    \caption{Efficiency analysis on Ambi-RVOS.}
    \label{tab:latency}
    \setlength{\tabcolsep}{5pt}
    \footnotesize
    \begin{tabular}{c|ccc|c}
      \toprule
       Method & $\mathcal{J}\&\mathcal{F}$ & Turns & Time & Train cost \\
    \midrule
    \rowcolor{gray!7.5}
    Baseline & 25.1 & - & 0.9 & - \\
    \midrule
    \rowcolor{rowhighlight}
    IC-Seg & \textbf{52.0} & \textbf{2.5} & 2.2 & 333.5 \\
    w/o step & 48.9 & 2.8 & 2.4 & 295.0 \\
    w/o step \& turn & 46.4 & 3.8 & 3.0 & 221.0 \\
    \bottomrule
    \end{tabular}
  \end{minipage}%
  \hfill
  \begin{minipage}[t]{0.49\textwidth}
    \centering
    \caption{User study on the reliability of the user simulator during test.}
    \label{tab:user_study}
    \setlength{\tabcolsep}{5pt}
    \footnotesize
    \begin{tabular}{c|c|ccc}
      \toprule
       Method & $\pi_B$ & $\mathcal{J}\&\mathcal{F}$ & $\mathcal{J}$ & $\mathcal{F}$ \\
      \midrule
      Qwen3-VL* & User & 33.9 & 32.1 & 35.8 \\
      IC-Seg & User & 55.1 & 52.1 & 58.1 \\
      \rowcolor{rowhighlight}
      IC-Seg & Qwen3-VL-32B & 53.6 & 50.4 & 56.8 \\
      \bottomrule
    \end{tabular}
  \end{minipage}%
  \vspace{-1em}
\end{table}

\subsection{Ablation Study}
Here, we conduct ablation studies with IC-Seg-4B to verify the individual component designs. 

\textbf{Effectiveness of Hi-GRPO.}
Table~\ref{tab:hi-grpo} validates the effectiveness of our three-level Hi-GRPO algorithm. The ``Baseline'' using solely the trajectory-level optimization yields 46.4 $\mathcal{J}\&\mathcal{F}$. Integrating either turn-level reward ($\mathcal{R}_{ent}$ or $\mathcal{R}_{eff}$) individually is effective, and combining them further raises the score to 48.9 $\mathcal{J}\&\mathcal{F}$. Finally, introducing the step-level supervision ($\mathcal{R}_{step}$) further boosts the performance to the optimal 52.0 $\mathcal{J}\&\mathcal{F}$, confirming the necessity of the complete hierarchical framework.

\textbf{Design of the Privileged Context.}
Table~\ref{tab:feedback} compares different privileged guidance $r$. ``Empty'' re-evaluate the rollout without guidance. ``Answer'' provides the ground-truth target and improves over ``Empty'' only marginally. In contrast, our expert ``Guidance'' achieves the highest accuracy by providing diagnostic, process-oriented feedback that identifies flaws in reasoning. We also evaluate SDPO~\cite{hubotter2026reinforcement}, which directly distills the privileged distribution without explicit reward optimization. Even when equipped with our expert guidance, its poor performance suggests that step-level distillation alone is insufficient for long-horizon interaction.

\textbf{Efficiency Analysis.}
Table~\ref{tab:latency} reports the efficiency on Ambi-RVOS, where ``Time'' denotes per-sample inference time and ``Train cost'' denotes per-step training time~(seconds). The ``Baseline'' denotes Qwen3-VL alone and without the user simulator, which is faster but obtains the worst performance. IC-Seg achieves the best accuracy with 2.5 clarification turns on average. Removing the step-level or turn-level design reduces accuracy and increases interaction turns, showing that Hi-GRPO improves both clarification quality and efficiency. Regarding training, IC-Seg's overhead is manageable: compared with trajectory-reward RL, the extra cost mainly comes from simulator calls and one additional forward pass. The latter bypasses text generation, increasing step time by only $\sim 13\%$ over w/o step level. Detailed per-difficulty results are provided in Appendix~\ref{sec:efficiency_analysis}.


\begin{figure}[t]
  \centering
  \includegraphics[width=.95\textwidth]{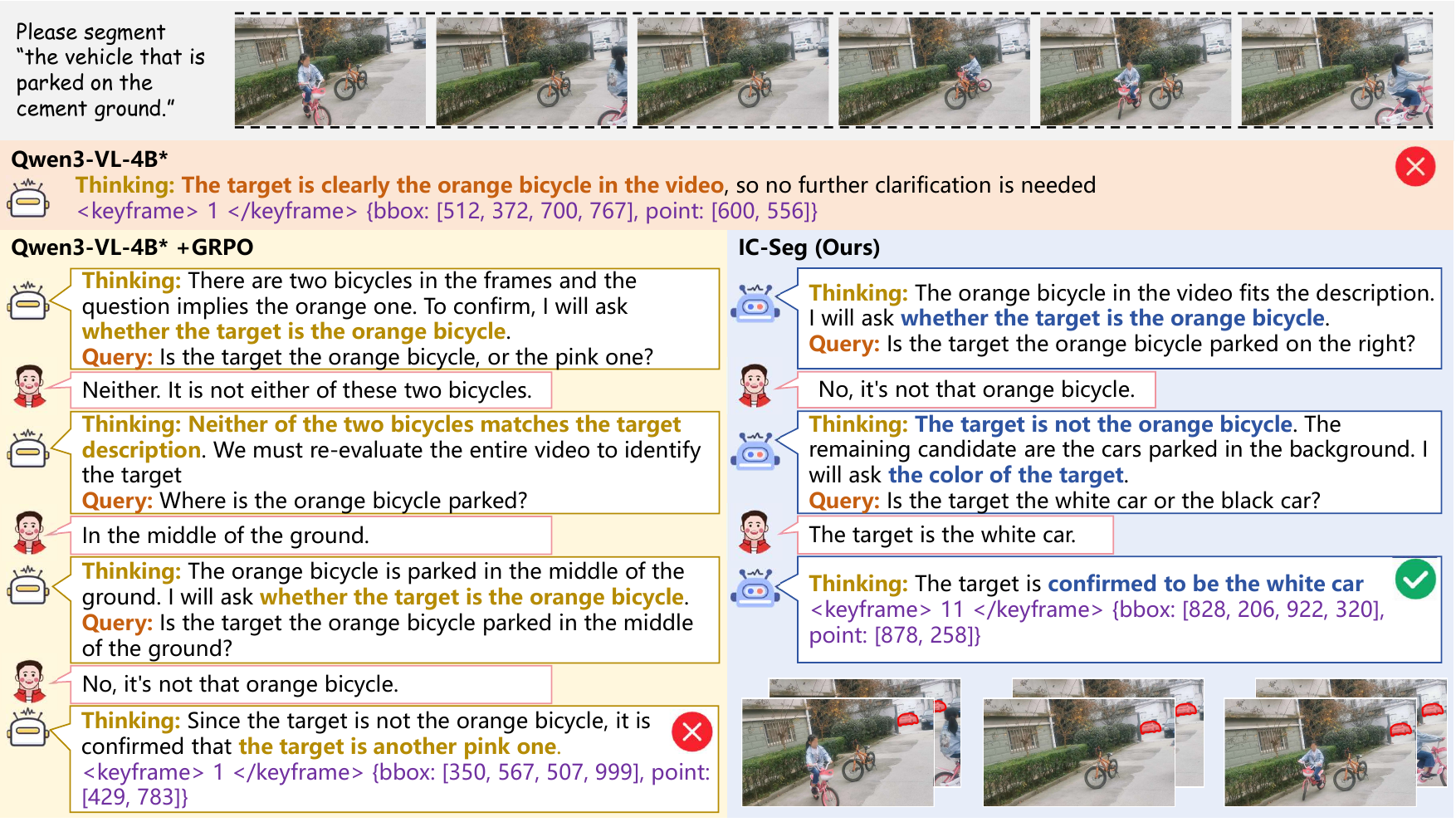}
  \caption{Qualitative comparisons among our IC-Seg and two baselines.}
  \label{fig:visual}
  \vspace{-1em}
\end{figure}

\begin{table}[t]
  \centering
  \begin{minipage}[t]{0.48\textwidth}
    \centering
    \caption{Impact of answering quality based on different size of user simulators during training.}
    \label{tab:user}
    \setlength{\tabcolsep}{11pt}
    \footnotesize
    \begin{tabular}{c|ccc}
      \toprule
       User Simulator & $\mathcal{J}\&\mathcal{F}$ & $\mathcal{J}$ & $\mathcal{F}$ \\
      \midrule
      Qwen3-VL-4B & 45.3 & 42.1 & 48.5 \\
      Qwen3-VL-8B & 47.5 & 44.4 & 50.7 \\
      \rowcolor{rowhighlight}
      Qwen3-VL-32B & 52.0 & 49.2 & 54.9 \\
      \bottomrule
    \end{tabular}
  \end{minipage}
  \hfill
  \begin{minipage}[t]{0.49\textwidth}
    \centering
    \caption{Impact of different user simulators during inference.}
    \label{tab:user_test}
    \setlength{\tabcolsep}{11pt}
    \footnotesize
    \begin{tabular}{c|ccc}
      \toprule
       User Simulator & $\mathcal{J}\&\mathcal{F}$ & $\mathcal{J}$ & $\mathcal{F}$ \\
      \midrule
      Qwen3-VL-8B & 47.9 & 44.9 & 50.8 \\
      \rowcolor{rowhighlight}
      Qwen3-VL-32B & 52.0 & 49.2 & 54.9 \\
      InternVL3.5-38B & 51.3 & 48.5 & 54.2 \\
      \bottomrule
    \end{tabular}
  \end{minipage}
  \vspace{-.5em}
\end{table}


\subsection{Analysis of User Simulator}
Since Ambi-RVOS relies on interaction during evaluation and training, the choice of user simulator may affect both measurement reliability and policy learning. We therefore analyze its reliability, training influence, and inference influence. Qwen3-VL-4B is used as the base model in this section.

\textbf{Evaluation Reliability.}
We first verify whether the user simulator provides a faithful approximation to real users. We randomly select 150 test samples and ask five human users to interact with the model independently, using the same information provided to the simulator. Table~\ref{tab:user_study} reports the average performance over these users. IC-Seg achieves similar results with the simulator and human users, supporting the reliability of simulator-based evaluation. Moreover, IC-Seg substantially outperforms Qwen3-VL* under real-user interaction, validating the effectiveness of our Hi-GRPO.

\textbf{Influence of Different User Simulators.}
We further study how user simulator quality affects IC-Seg during training and inference. In Table~\ref{tab:user}, the inference simulator is fixed as Qwen3-VL-32B while the training simulator varies. Performance steadily improves from 45.3 to 52.0 as the training simulator scales from 4B to 32B, indicating that stronger simulators provide more stable and informative feedback during training. In Table~\ref{tab:user_test}, we fix training with Qwen3-VL-32B and vary the inference simulator. A stronger inference simulator improves performance by answering more reliably. Meanwhile, InternVL3.5-38B achieves comparable results to Qwen3-VL-32B, suggesting that IC-Seg does not simply overfit to one simulator family.

\subsection{Qualitative Results}
We present a representative ambiguous case in \Cref{fig:visual}, where the query can refer to multiple static vehicles. Qwen3-VL-4B* directly guesses the orange bicycle and produces an incorrect mask. Qwen3-VL-4B*+\textit{GRPO} denotes the model trained with only our trajectory-level optimization, which attempts clarification but repeatedly focuses on the same orange bicycle after negative feedback and finally falls back to another unsupported guess. In contrast, IC-Seg uses the user's responses to exclude implausible candidates and then verifies the white car. This example shows that Hi-GRPO enables more effective use of multi-turn feedback for intent clarification.

\section{Conclusion}
In this paper, we study intent clarification in language-guided segmentation via IC-Seg, an agentic framework that resolves ambiguity through user interaction. To train this behavior, we proposed Hi-GRPO, a hierarchical RL strategy that combines trajectory, turn and step levels optimization for dense and effective supervision. We further established Ambi-RVOS, a benchmark for Referring Video Object Segmentation that contains Ambiguous user queries. Experiments on Ambi-RVOS and existing reasoning segmentation benchmarks demonstrate that IC-Seg effectively resolves ambiguous references while preserving strong grounding and segmentation capabilities. We hope this work encourages future research to move language-guided segmentation beyond passive grounding toward more interactive and intent-aware visual systems.

\newpage
\bibliographystyle{plainnat}
\bibliography{neurips_2026}

@String(CVPR= {IEEE Conf. Comput. Vis. Pattern Recog.})

@String(ECCV= {Eur. Conf. Comput. Vis.})

@String(AAAI = {AAAI})

@String(CVPR  = {CVPR})

@String(ECCV  = {ECCV})

@article{shao2024deepseekmath,
  title={Deepseekmath: Pushing the limits of mathematical reasoning in open language models, 2024},
  author={Shao, Zhihong and Wang, Peiyi and Zhu, Qihao and Xu, Runxin and Song, Junxiao and Bi, Xiao and Zhang, Haowei and Zhang, Mingchuan and Li, YK and Wu, Yang and others},
  journal={URL https://arxiv. org/abs/2402.03300},
  volume={2},
  number={3},
  pages={5},
  year={2024}
}

@inproceedings{yan2024visa,
  title={Visa: Reasoning video object segmentation via large language models},
  author={Yan, Cilin and Wang, Haochen and Yan, Shilin and Jiang, Xiaolong and Hu, Yao and Kang, Guoliang and Xie, Weidi and Gavves, Efstratios},
  booktitle={European Conference on Computer Vision (ECCV)},
  pages={98--115},
  year={2024},
  organization={Springer}
}

@article{bai2024one,
  title={One token to seg them all: Language instructed reasoning segmentation in videos},
  author={Bai, Zechen and He, Tong and Mei, Haiyang and Wang, Pichao and Gao, Ziteng and Chen, Joya and Zhang, Zheng and Shou, Mike Zheng},
  journal={Advances in Neural Information Processing Systems},
  volume={37},
  pages={6833--6859},
  year={2024}
}

@article{kao2025cot,
  title={CoT-RVS: Zero-Shot Chain-of-Thought Reasoning Segmentation for Videos},
  author={Kao, Shiu-hong and Tai, Yu-Wing and Tang, Chi-Keung},
  journal={arXiv preprint arXiv:2505.18561},
  year={2025}
}

@inproceedings{lin2025glus,
  title={GLUS: Global-Local Reasoning Unified into A Single Large Language Model for Video Segmentation},
  author={Lin, Lang and Yu, Xueyang and Pang, Ziqi and Wang, Yu-Xiong},
  booktitle={Proceedings of the IEEE/CVF Conference on Computer Vision and Pattern Recognition},
  year={2025}
}

@inproceedings{liang2025referdino,
    title={ReferDINO: Referring Video Object Segmentation with Visual Grounding Foundations},
    author={Liang, Tianming and Lin, Kun-Yu and Tan, Chaolei and Zhang, Jianguo and Zheng, Wei-Shi and Hu, Jian-Fang},
    booktitle={Proceedings of the IEEE/CVF International Conference on Computer Vision},
    year={2025}
}

@InProceedings{cuttano2025samwise,
    author    = {Cuttano, Claudia and Trivigno, Gabriele and Rosi, Gabriele and Masone, Carlo and Averta, Giuseppe},
    title     = {SAMWISE: Infusing Wisdom in SAM2 for Text-Driven Video Segmentation},
    booktitle = {Proceedings of the Computer Vision and Pattern Recognition Conference (CVPR)},
    month     = {June},
    year      = {2025},
    pages     = {3395-3405}
}

@inproceedings{ding2023mevis,
  title={MeViS: A large-scale benchmark for video segmentation with motion expressions},
  author={Ding, Henghui and Liu, Chang and He, Shuting and Jiang, Xudong and Loy, Chen Change},
  booktitle={Proceedings of the IEEE/CVF international conference on computer vision},
  pages={2694--2703},
  year={2023}
}

@inproceedings{seo2020urvos,
  title={Urvos: Unified referring video object segmentation network with a large-scale benchmark},
  author={Seo, Seonguk and Lee, Joon-Young and Han, Bohyung},
  booktitle={European conference on computer vision},
  pages={208--223},
  year={2020},
  organization={Springer}
}

@inproceedings{botach2022end,
  title={End-to-end referring video object segmentation with multimodal transformers},
  author={Botach, Adam and Zheltonozhskii, Evgenii and Baskin, Chaim},
  booktitle={Proceedings of the IEEE/CVF Conference on Computer Vision and Pattern Recognition},
  pages={4985--4995},
  year={2022}
}

@article{liang2025longrvos,
  title={Long-RVOS: A Comprehensive Benchmark for Long-term Referring Video Object Segmentation},
  author={Liang, Tianming and Jiang, Haichao and Yang, Yuting and Tan, Chaolei and Li, Shuai and Zheng, Wei-Shi and Hu, Jian-Fang},
  journal={arXiv preprint arXiv:2505.12702},
  year={2025}
}

@inproceedings{liu2025unipixel,
  title={UniPixel: Unified Object Referring and Segmentation for Pixel-Level Visual Reasoning},
  author={Liu, Ye and Ma, Zongyang and Pu, Junfu and Qi, Zhongang and Wu, Yang and Ying, Shan and Chen, Chang Wen},
  booktitle={Advances in Neural Information Processing Systems (NeurIPS)},
  year={2025}
}

@article{jiang2026refer,
  title={Refer-Agent: A Collaborative Multi-Agent System with Reasoning and Reflection for Referring Video Object Segmentation},
  author={Jiang, Haichao and Liang, Tianming and Zheng, Wei-Shi and Hu, Jian-Fang},
  journal={arXiv preprint arXiv:2602.03595},
  year={2026}
}

@inproceedings{xu2026videoseg,
  title={Videoseg-r1: reasoning video object segmentation via reinforcement learning},
  author={Xu, Zishan and Guo, Yifu and Lu, Yuquan and Yang, Fengyu and Li, Junxin and Cai, Lihua},
  booktitle={Proceedings of the AAAI Conference on Artificial Intelligence},
  volume={40},
  number={14},
  pages={11496--11504},
  year={2026}
}

@inproceedings{gong2026veason,
  title={Reinforcing Video Reasoning Segmentation to Think Before It Segments},
  author={Gong, Sitong and Zhuge, Yunzhi and Zhang, Lu and Yu, Jiazuo and Jia, Xu and Zhang, Pingping and Lu, Huchuan},
  booktitle={Proceedings of the IEEE/CVF Conference on Computer Vision and Pattern Recognition},
  year={2026}
}

@article{liang2026seg,
  title={Seg-ReSearch: Segmentation with Interleaved Reasoning and External Search},
  author={Liang, Tianming and Du, Qirui and Hu, Jian-Fang and Jiang, Haichao and Lin, Zicheng and Zheng, Wei-Shi},
  journal={arXiv preprint arXiv:2602.04454},
  year={2026}
}

@inproceedings{carion2020end,
  title={End-to-end object detection with transformers},
  author={Carion, Nicolas and Massa, Francisco and Synnaeve, Gabriel and Usunier, Nicolas and Kirillov, Alexander and Zagoruyko, Sergey},
  booktitle={European conference on computer vision},
  pages={213--229},
  year={2020},
  organization={Springer}
}

@article{luo2023soc,
  title={Soc: Semantic-assisted object cluster for referring video object segmentation},
  author={Luo, Zhuoyan and Xiao, Yicheng and Liu, Yong and Li, Shuyan and Wang, Yitong and Tang, Yansong and Li, Xiu and Yang, Yujiu},
  journal={Advances in Neural Information Processing Systems},
  volume={36},
  pages={26425--26437},
  year={2023}
}

@inproceedings{miao2023spectrum,
  title={Spectrum-guided multi-granularity referring video object segmentation},
  author={Miao, Bo and Bennamoun, Mohammed and Gao, Yongsheng and Mian, Ajmal},
  booktitle={Proceedings of the IEEE/CVF International Conference on Computer Vision},
  pages={920--930},
  year={2023}
}

@article{ravi2024sam,
  title={Sam 2: Segment anything in images and videos},
  author={Ravi, Nikhila and Gabeur, Valentin and Hu, Yuan-Ting and Hu, Ronghang and Ryali, Chaitanya and Ma, Tengyu and Khedr, Haitham and R{\"a}dle, Roman and Rolland, Chloe and Gustafson, Laura and others},
  journal={arXiv preprint arXiv:2408.00714},
  year={2024}
}

@inproceedings{lai2024lisa,
  title={Lisa: Reasoning segmentation via large language model},
  author={Lai, Xin and Tian, Zhuotao and Chen, Yukang and Li, Yanwei and Yuan, Yuhui and Liu, Shu and Jia, Jiaya},
  booktitle={Proceedings of the IEEE/CVF Conference on Computer Vision and Pattern Recognition},
  pages={9579--9589},
  year={2024}
}

@inproceedings{wang2025object,
  title={Object-centric video question answering with visual grounding and referring},
  author={Wang, Haochen and Chen, Qirui and Yan, Cilin and Cai, Jiayin and Jiang, Xiaolong and Hu, Yao and Xie, Weidi and Gavves, Stratis},
  booktitle={Proceedings of the IEEE/CVF International Conference on Computer Vision},
  pages={22274--22284},
  year={2025}
}

@article{liu2025seg,
  title={Seg-zero: Reasoning-chain guided segmentation via cognitive reinforcement},
  author={Liu, Yuqi and Peng, Bohao and Zhong, Zhisheng and Yue, Zihao and Lu, Fanbin and Yu, Bei and Jia, Jiaya},
  journal={arXiv preprint arXiv:2503.06520},
  year={2025}
}

@article{li2026rethinking,
  title={Rethinking On-Policy Distillation of Large Language Models: Phenomenology, Mechanism, and Recipe},
  author={Li, Yaxuan and Zuo, Yuxin and He, Bingxiang and Zhang, Jinqian and Xiao, Chaojun and Qian, Cheng and Yu, Tianyu and Gao, Huan-ang and Yang, Wenkai and Liu, Zhiyuan and others},
  journal={arXiv preprint arXiv:2604.13016},
  year={2026}
}

@article{hubotter2026reinforcement,
  title={Reinforcement Learning via Self-Distillation},
  author={H{\"u}botter, Jonas and L{\"u}beck, Frederike and Behric, Lejs and Baumann, Anton and Bagatella, Marco and Marta, Daniel and Hakimi, Ido and Shenfeld, Idan and Buening, Thomas Kleine and Guestrin, Carlos and others},
  journal={arXiv preprint arXiv:2601.20802},
  year={2026}
}

@article{yang2026self,
  title={Self-Distilled RLVR},
  author={Yang, Chenxu and Qin, Chuanyu and Si, Qingyi and Chen, Minghui and Gu, Naibin and Yao, Dingyu and Lin, Zheng and Wang, Weiping and Wang, Jiaqi and Duan, Nan},
  journal={arXiv preprint arXiv:2604.03128},
  year={2026}
}

@misc{carion2025sam3,
      title={SAM 3: Segment Anything with Concepts},
      author={Nicolas Carion and Laura Gustafson and Yuan-Ting Hu and Shoubhik Debnath and Ronghang Hu and Didac Suris and Chaitanya Ryali and Kalyan Vasudev Alwala and Haitham Khedr and Andrew Huang and Jie Lei and Tengyu Ma and Baishan Guo and Arpit Kalla and Markus Marks and Joseph Greer and Meng Wang and Peize Sun and Roman Rädle and Triantafyllos Afouras and Effrosyni Mavroudi and Katherine Xu and Tsung-Han Wu and Yu Zhou and Liliane Momeni and Rishi Hazra and Shuangrui Ding and Sagar Vaze and Francois Porcher and Feng Li and Siyuan Li and Aishwarya Kamath and Ho Kei Cheng and Piotr Dollár and Nikhila Ravi and Kate Saenko and Pengchuan Zhang and Christoph Feichtenhofer},
      year={2025},
      eprint={2511.16719},
      archivePrefix={arXiv},
      primaryClass={cs.CV},
      url={https://arxiv.org/abs/2511.16719},
}

@article{feng2025onethinker,
  title={Onethinker: All-in-one reasoning model for image and video},
  author={Feng, Kaituo and Zhang, Manyuan and Li, Hongyu and Fan, Kaixuan and Chen, Shuang and Jiang, Yilei and Zheng, Dian and Sun, Peiwen and Zhang, Yiyuan and Sun, Haoze and others},
  journal={arXiv preprint arXiv:2512.03043},
  year={2025}
}

@inproceedings{huang2025samr,
  title={{SAM}-R1: Leveraging {SAM} for Reward Feedback in Multimodal Segmentation via Reinforcement Learning},
  author={Jiaqi Huang and Zunnan Xu and Jun Zhou and Ting Liu and Yicheng Xiao and Mingwen Ou and Bowen Ji and Xiu Li and Kehong Yuan},
  booktitle={The Thirty-ninth Annual Conference on Neural Information Processing Systems (NeurIPS)},
  year={2025}
}

@article{rafailov2023direct,
  title={Direct preference optimization: Your language model is secretly a reward model},
  author={Rafailov, Rafael and Sharma, Archit and Mitchell, Eric and Manning, Christopher D and Ermon, Stefano and Finn, Chelsea},
  journal={Advances in neural information processing systems},
  volume={36},
  pages={53728--53741},
  year={2023}
}

@inproceedings{yu2025dapo,
title={{DAPO}: An Open-Source {LLM} Reinforcement Learning System at Scale},
author={Qiying Yu and Zheng Zhang and Ruofei Zhu and Yufeng Yuan and Xiaochen Zuo and YuYue and Weinan Dai and Tiantian Fan and Gaohong Liu and Juncai Liu and LingJun Liu and Xin Liu and Haibin Lin and Zhiqi Lin and Bole Ma and Guangming Sheng and Yuxuan Tong and Chi Zhang and Mofan Zhang and Ru Zhang and Wang Zhang and Hang Zhu and Jinhua Zhu and Jiaze Chen and Jiangjie Chen and Chengyi Wang and Hongli Yu and Yuxuan Song and Xiangpeng Wei and Hao Zhou and Jingjing Liu and Wei-Ying Ma and Ya-Qin Zhang and Lin Yan and Yonghui Wu and Mingxuan Wang},
booktitle={The Thirty-ninth Annual Conference on Neural Information Processing Systems (NeurIPS)},
year={2025}
}

@article{bai2025qwen3vl,
  title={Qwen3-VL Technical Report},
  author={Bai, Shuai and Cai, Yuxuan and Chen, Ruizhe and Chen, Keqin and Chen, Xionghui and Cheng, Zesen and Deng, Lianghao and Ding, Wei and Gao, Chang and Ge, Chunjiang and Ge, Wenbin and Guo, Zhifang and Huang, Qidong and Huang, Jie and Huang, Fei and Hui, Binyuan and Jiang, Shutong and Li, Zhaohai and Li, Mingsheng and Li, Mei and Li, Kaixin and Lin, Zicheng and Lin, Junyang and Liu, Xuejing and Liu, Jiawei and Liu, Chenglong and Liu, Yang and Liu, Dayiheng and Liu, Shixuan and Lu, Dunjie and Luo, Ruilin and Lv, Chenxu and Men, Rui and Meng, Lingchen and Ren, Xuancheng and Ren, Xingzhang offense and Song, Sibo and Sun, Yuchong and Tang, Jun and Tu, Jianhong and Wan, Jianqiang and Wang, Peng and Wang, Pengfei and Wang, Qiuyue and Wang, Yuxuan and Xie, Tianbao and Xu, Yiheng and Xu, Haiyang and Xu, Jin and Yang, Zhibo and Yang, Mingkun and Yang, Jianxin and Yang, An and Yu, Bowen and Zhang, Fei and Zhang, Hang and Zhang, Xi and Zheng, Bo and Zhong, Humen and Zhou, Jingren and Zhou, Fan and Zhou, Jing and Zhu, Yuanzhi and Zhu, Ke},
  journal={arXiv preprint arXiv:2511.21631},
  year={2025}
}

@inproceedings{jin2025searchr,
  title={Search-R1: Training {LLM}s to Reason and Leverage Search Engines with Reinforcement Learning},
  author={Bowen Jin and Hansi Zeng and Zhenrui Yue and Jinsung Yoon and Sercan O Arik and Dong Wang and Hamed Zamani and Jiawei Han},
  booktitle={Second Conference on Language Modeling},
  year={2025}
}

@inproceedings{zheng2025stepsearch,
  title={StepSearch: Igniting LLMs Search Ability via Step-Wise Proximal Policy Optimization},
  author={Zheng, Xuhui and An, Kang and Wang, Ziliang and Wang, Yuhang and Wu, Yichao},
  booktitle={Proceedings of the 2025 Conference on Empirical Methods in Natural Language Processing (EMNLP)},
  pages={21805--21830},
  year={2025}
}

@article{geng2025webwatcher,
  title={Webwatcher: Breaking new frontier of vision-language deep research agent},
  author={Geng, Xinyu and Xia, Peng and Zhang, Zhen and Wang, Xinyu and Wang, Qiuchen and Ding, Ruixue and Wang, Chenxi and Wu, Jialong and Zhao, Yida and Li, Kuan and others},
  journal={arXiv preprint arXiv:2508.05748},
  year={2025}
}

@article{yang2025qwen3,
  title={Qwen3 technical report},
  author={Yang, An and Li, Anfeng and Yang, Baosong and Zhang, Beichen and Hui, Binyuan and Zheng, Bo and Yu, Bowen and Gao, Chang and Huang, Chengen and Lv, Chenxu and others},
  journal={arXiv preprint arXiv:2505.09388},
  year={2025}
}

@inproceedings{agarwal2024policy,
  title={On-policy distillation of language models: Learning from self-generated mistakes},
  author={Agarwal, Rishabh and Vieillard, Nino and Zhou, Yongchao and Stanczyk, Piotr and Garea, Sabela Ramos and Geist, Matthieu and Bachem, Olivier},
  booktitle={The twelfth international conference on learning representations},
  year={2024}
}

@inproceedings{das2017visual,
  title={Visual dialog},
  author={Das, Abhishek and Kottur, Satwik and Gupta, Khushi and Singh, Avi and Yadav, Deshraj and Moura, Jos{\'e} MF and Parikh, Devi and Batra, Dhruv},
  booktitle={Proceedings of the IEEE conference on computer vision and pattern recognition},
  pages={326--335},
  year={2017}
}

@inproceedings{de2017guesswhat,
  title={Guesswhat?! visual object discovery through multi-modal dialogue},
  author={De Vries, Harm and Strub, Florian and Chandar, Sarath and Pietquin, Olivier and Larochelle, Hugo and Courville, Aaron},
  booktitle={Proceedings of the IEEE Conference on Computer Vision and Pattern Recognition},
  pages={5503--5512},
  year={2017}
}

@inproceedings{wu2021fashion,
  title={Fashion iq: A new dataset towards retrieving images by natural language feedback},
  author={Wu, Hui and Gao, Yupeng and Guo, Xiaoxiao and Al-Halah, Ziad and Rennie, Steven and Grauman, Kristen and Feris, Rogerio},
  booktitle={Proceedings of the IEEE/CVF Conference on computer vision and pattern recognition},
  pages={11307--11317},
  year={2021}
}

@article{liu2023visual,
  title={Visual instruction tuning},
  author={Liu, Haotian and Li, Chunyuan and Wu, Qingyang and Lee, Yong Jae},
  journal={Advances in neural information processing systems},
  volume={36},
  pages={34892--34916},
  year={2023}
}

@article{chen2023llava,
  title={Llava-interactive: An all-in-one demo for image chat, segmentation, generation and editing},
  author={Chen, Wei-Ge and Spiridonova, Irina and Yang, Jianwei and Gao, Jianfeng and Li, Chunyuan},
  journal={arXiv preprint arXiv:2311.00571},
  year={2023}
}

@inproceedings{ni2026survey,
  title={A Survey on LLM-based Conversational User Simulation},
  author={Ni, Bo and Wang, Yu and Wang, Leyao and Kveton, Branislav and Dernoncourt, Franck and Xia, Yu and Chen, Hongjie and Luera, Reuben and Basu, Samyadeep and Mukherjee, Subhojyoti and others},
  booktitle={Proceedings of the 19th Conference of the European Chapter of the Association for Computational Linguistics (Volume 1: Long Papers)},
  pages={4266--4301},
  year={2026}
}

@inproceedings{Li2025IAGIB,
  title={IAG: Input-aware Backdoor Attack on VLM-based Visual Grounding},
  author={Junxian Li and Beining Xu and Di Zhang},
  year={2025},
  url={https://api.semanticscholar.org/CorpusID:280641739}
}

@article{Munasinghe2024VideoGLaMMA,
  title={VideoGLaMM : A Large Multimodal Model for Pixel-Level Visual Grounding in Videos},
  author={Shehan Munasinghe and Hanan Gani and Wenqi Zhu and Jiale Cao and Eric P. Xing and Fahad Shahbaz Khan and Salman H. Khan},
  journal={2025 IEEE/CVF Conference on Computer Vision and Pattern Recognition (CVPR)},
  year={2024},
  pages={19036-19046},
  url={https://api.semanticscholar.org/CorpusID:273878153}
}

\newpage
\appendix

\section{More Implementation Details}
\label{sec:more_details}
We implement our method based on the VERL framework. We set the maximum query turns to 5 in both training and inference stages. All experiments are conducted on a single node with 8 NVIDIA RTX A6000 GPUs.
For video segmentation settings, we uniformly sample 6 frames from the video and all input frames are resized to $448 \times 448$, while the later selected keyframe by our model is resized to $864 \times 864$. For image segmentation settings, input frame is resized to $864 \times 864$.
For both IC-Seg-4B and IC-Seg-8B, we train the policy for 100 optimization steps with GRPO-style group sampling. The learning rate is set to $1\times10^{-6}$ with 10 warmup steps. We disable additional KL and entropy regularization.
The maximum prompt length, action length, observation length, and response length are set to 2176, 4096, 2048, and 6144 tokens, respectively. We mask tool observations in the policy loss so that the model is optimized only on its own generated actions and responses.
Following RLSD, token re-evaluation uses a frozen policy copy that is synchronized with the current policy every 10 training steps, which improves the stability of token-level guidance. 
When re-evaluating, we concatenate the expert guidance $r$ with the original prompt and previous response to obtain a new input. Below shows a simple example.

\begin{tcolorbox}[
    colback=white,
    colframe=gray!85,
    coltitle=white,
    fonttitle=\bfseries,
    title={Example: Concatenation input with expert guidance $r$},
    boxrule=0.8pt,
    arc=3pt,
    left=0pt, right=0pt, top=5pt, bottom=5pt,
    label={box:example_guidance}
]
\small
\textbf{Original Prompt:} Your role is a video target identification assistant...Please find `A lamb chop being roasted'.

\vspace{0.2em}

\textbf{A priori guidance:} While identifying the cooking state effectively narrows candidates, further refinement is required. To isolate the target, ask about precise spatial positions. For localization, select points directly on the meat surface. Frame 5 is an unsuitable keyframe as the chop is being flipped, causing severe occlusion.

\vspace{0.2em}

\textbf{Use the guidance to help you judge the previous attempt.}

\vspace{0.2em}

\textbf{Previous Response:} The frames show multiple lamb chops, each with a similar structure...
\end{tcolorbox}

\section{Algorithmic Details of Hi-GRPO}
\label{sec:code}
Algorithm~\ref{alg:hi-grpo} summarizes the overall training procedure of Hi-GRPO: 1) sample on-policy multi-turn trajectories, 2) compute sequence-level advantages from trajectory and turn rewards, 3) finally refine token-level credit assignment using diagnostic expert guidance before the GRPO update.

\begin{algorithm}[h]
\caption{Hi-GRPO: Hierarchical Group Relative Policy Optimization}
\label{alg:hi-grpo}
\begin{algorithmic}[1]
\Require Initial policy $\pi_{\theta_{\text{old}}}$, user simulator $\pi_B$, dataset $\mathcal{S} = \{x_j\}_{j=1}^N$ (each $x$ contains video and query), group size $G$, reward weight $\alpha$, mixing coefficient $\lambda$, clip bound $\epsilon_f$
\For{each training iteration}
    \State Sample a batch of multi-modal inputs $\{x\}$ from $\mathcal{S}$
    \For{each input $x$}
        \State Sample $G$ trajectories $\{y_i\}_{i=1}^G$ from $\pi_{\theta_{\text{old}}}$ and record dialog histories $\{h_i\}_{i=1}^G$ with $\pi_B$
        
        \For{$i = 1, \dots, G$}
            \State Compute total sequence reward $\mathcal{R}_i = \alpha \cdot \mathcal{R}_{i}^{\text{turn}} + \mathcal{R}_{i}^{\text{traj}}$
        \EndFor
        \State Compute $A_i = \frac{\mathcal{R}_i - \mu_G}{\sigma_G}$, where $\mu_G, \sigma_G$ are the mean and std of $\{\mathcal{R}_1, \dots, \mathcal{R}_G\}$
        
        \For{$i = 1, \dots, G$}
            \State Generate expert guidance $r_i$ using $\pi_B(x, h_i, y_i, \text{ground-truth})$
            \For{$t = 1, \dots, |y_i|$}
                \State Compute probability ratio $f_{i,t} \gets \texttt{sg}\!\left( \frac{\pi_\theta(y_{i,t} \mid x, r_i, y_{i,<t})}{\pi_\theta(y_{i,t} \mid x, y_{i,<t})} \right)$ \Comment{\texttt{sg}: stop gradient}
                \State $\tilde{A}_{i,t} \gets A_i \left[ (1 - \lambda) + \lambda \cdot \text{clip}\left(f_{i,t}^{\text{sign}(A_i)},\, 1 - \epsilon_f,\, 1 + \epsilon_f\right) \right]$
            \EndFor
        \EndFor
    \EndFor
    
    \State Update $\theta$ by maximizing $\mathcal{L}(\theta) = \frac{1}{G} \sum_{i=1}^{G} \frac{1}{|y_i|} \sum_{t=1}^{|y_i|} \frac{\pi_\theta(y_{i,t} \mid x, y_{i,<t})}{\pi_{\theta_{\text{old}}}(y_{i,t} \mid x, y_{i,<t})} \tilde{A}_{i,t}$
    \State $\pi_{\theta_{\text{old}}} \gets \pi_\theta$
\EndFor
\end{algorithmic}
\end{algorithm}

\section{Training Dynamics of IC-Seg}
\begin{figure}[h]
  \centering
  \includegraphics[width=1.0\textwidth]{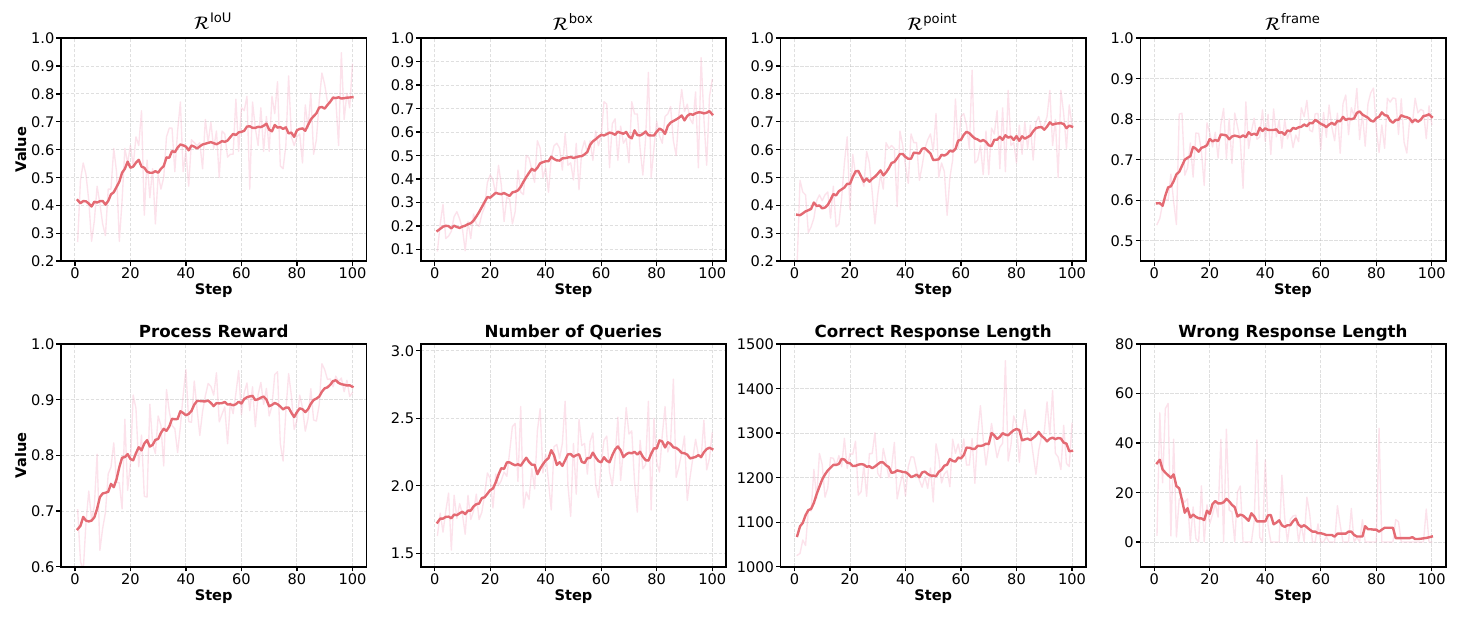}
  \caption{Training Dynamics of IC-Seg-8B.}
  \label{fig:dynamics}
\end{figure}

Figure~\ref{fig:dynamics} shows the training dynamics of IC-Seg-8B. The localization-related rewards, including $\mathcal{R}^{\text{IoU}}$, $\mathcal{R}^{\text{box}}$, $\mathcal{R}^{\text{point}}$, and $\mathcal{R}^{\text{frame}}$, steadily increase during training, indicating that the model gradually improves its final grounding accuracy and keyframe selection. The process reward also rises consistently, suggesting that Hi-GRPO encourages more effective clarification behavior rather than only optimizing the final segmentation output.
We further observe that the number of queries increases at the early stage and then stabilizes around a moderate range, showing that the model learns to ask additional questions when needed without entering excessive interaction loops. Meanwhile, the length of wrong responses quickly decreases, while correct responses remain sufficiently detailed. These trends suggest that training suppresses invalid or unproductive trajectories and improves the model's ability to conduct concise yet informative intent clarification.

\begin{table*}[b]
\setlength\tabcolsep{3.5pt}
\footnotesize
\centering
\caption{Comparison with SFT baselines on Ambi-RVOS. $\Delta$ rows denote changes from the corresponding untrained baseline after using the same 120 Ambi-RVOS training samples.}
\label{tab:sft_baselines}
\begin{tabular}{l|ccc|ccc|ccc|ccc}
\toprule
\multirow{2}{*}{Method} & \multicolumn{3}{c|}{\textbf{Simple}} & \multicolumn{3}{c|}{\textbf{Medium}} & \multicolumn{3}{c|}{\textbf{Difficult}} & \multicolumn{3}{c}{\textbf{Overall}} \\
\cmidrule(lr){2-13}
 & $\mathcal{J}\&\mathcal{F}$ & $\mathcal{J}$ & $\mathcal{F}$ & $\mathcal{J}\&\mathcal{F}$ & $\mathcal{J}$ & $\mathcal{F}$ & $\mathcal{J}\&\mathcal{F}$ & $\mathcal{J}$ & $\mathcal{F}$ & $\mathcal{J}\&\mathcal{F}$ & $\mathcal{J}$ & $\mathcal{F}$ \\
\midrule
GLUS-7B {\scriptsize\color{mydarkblue}CVPR'25} & 37.8 & 33.9 & 41.8 & 27.3 & 22.7 & 31.8 & 18.1 & 14.7 & 21.6 & 27.1 & 23.0 & 31.1 \\
GLUS-7B-SFT & 26.1 & 23.0 & 29.1 & 18.0 & 15.6 & 20.3 & 10.0 & 7.9 & 12.1 & 17.5 & 15.0 & 19.9 \\
$\Delta$ & \textcolor{mutedbluegray}{-11.7} & \textcolor{mutedbluegray}{-10.9} & \textcolor{mutedbluegray}{-12.7} & \textcolor{mutedbluegray}{-9.3} & \textcolor{mutedbluegray}{-7.1} & \textcolor{mutedbluegray}{-11.5} & \textcolor{mutedbluegray}{-8.1} & \textcolor{mutedbluegray}{-6.8} & \textcolor{mutedbluegray}{-9.5} & \textcolor{mutedbluegray}{-9.6} & \textcolor{mutedbluegray}{-8.0} & \textcolor{mutedbluegray}{-11.2} \\
\midrule
UniPixel-7B{\scriptsize\color{mydarkblue}NeurIPS'25} & 42.4 & 37.9 & 46.8 & 27.7 & 23.3 & 32.2 & 20.3 & 16.6 & 24.0 & 29.2 & 25.0 & 33.3 \\
UniPixel-7B-SFT & 37.8 & 34.6 & 40.9 & 25.0 & 22.3 & 27.7 & 18.9 & 16.8 & 21.0 & 26.4 & 23.8 & 29.0 \\
$\Delta$ & \textcolor{mutedbluegray}{-4.6} & \textcolor{mutedbluegray}{-3.3} & \textcolor{mutedbluegray}{-5.9} & \textcolor{mutedbluegray}{-2.7} & \textcolor{mutedbluegray}{-1.0} & \textcolor{mutedbluegray}{-4.5} & \textcolor{mutedbluegray}{-1.4} & \textcolor{PineGreen}{+0.2} & \textcolor{mutedbluegray}{-3.0} & \textcolor{mutedbluegray}{-2.8} & \textcolor{mutedbluegray}{-1.2} & \textcolor{mutedbluegray}{-4.3} \\
\midrule
Qwen3-VL-4B* & 35.8 & 33.5 & 38.0 & 31.6 & 29.8 & 33.5 & 24.5 & 23.1 & 26.0 & 30.4 & 28.6 & 32.2 \\
\rowcolor{rowhighlight}
IC-Seg-4B & \textbf{63.3} & \textbf{60.3} & \textbf{66.2} & \textbf{54.3} & \textbf{51.0} & \textbf{57.6} & \textbf{40.3} & \textbf{38.0} & \textbf{42.6} & \textbf{52.0} & \textbf{49.2} & \textbf{54.9} \\
$\Delta$ & \textcolor{PineGreen}{+27.5} & \textcolor{PineGreen}{+26.8} & \textcolor{PineGreen}{+28.2} & \textcolor{PineGreen}{+22.7} & \textcolor{PineGreen}{+21.2} & \textcolor{PineGreen}{+24.1} & \textcolor{PineGreen}{+15.8} & \textcolor{PineGreen}{+14.9} & \textcolor{PineGreen}{+16.6} & \textcolor{PineGreen}{+21.6} & \textcolor{PineGreen}{+20.6} & \textcolor{PineGreen}{+22.7} \\
\midrule
Qwen3-VL-8B* & 43.8 & 40.6 & 47.0 & 38.6 & 36.1 & 41.1 & 28.2 & 25.8 & 30.6 & 36.5 & 33.9 & 39.2 \\
\rowcolor{rowhighlight}
IC-Seg-8B & \textbf{63.7} & \textbf{60.3} & \textbf{67.0} & \textbf{57.4} & \textbf{54.1} & \textbf{60.8} & \textbf{45.5} & \textbf{42.6} & \textbf{48.3} & \textbf{55.1} & \textbf{51.9} & \textbf{58.3} \\
$\Delta$ & \textcolor{PineGreen}{+19.9} & \textcolor{PineGreen}{+19.7} & \textcolor{PineGreen}{+20.0} & \textcolor{PineGreen}{+18.8} & \textcolor{PineGreen}{+18.0} & \textcolor{PineGreen}{+19.7} & \textcolor{PineGreen}{+17.3} & \textcolor{PineGreen}{+16.8} & \textcolor{PineGreen}{+17.7} & \textcolor{PineGreen}{+18.6} & \textcolor{PineGreen}{+18.0} & \textcolor{PineGreen}{+19.1} \\
\bottomrule
\end{tabular}
\end{table*}

\section{Comparison with SFT Baselines}\label{sec:sft}
To verify whether the improvement of IC-Seg simply comes from the additional 120 training samples, we further fine-tune representative baselines on the same Ambi-RVOS training set.
Since GLUS and UniPixel are originally trained with supervised fine-tuning, we follow their training paradigm and adapt them to Ambi-RVOS using the same SFT protocol. As shown in Table~\ref{tab:sft_baselines}, directly fine-tuning them on the 120 training samples does not improve their robustness to ambiguous references. However, GLUS-7B drops by $9.6$ overall $\mathcal{J}\&\mathcal{F}$ after SFT, and UniPixel-7B also decreases by $2.8$ points. This suggests that SFT mainly encourages the model to memorize fixed supervision patterns, which is insufficient for Ambi-RVOS where success depends on exploratory questioning, interactive disambiguation, and deciding when enough evidence has been collected. In contrast, IC-Seg obtains large and consistent gains over its Qwen3-VL* initialization across all difficulty levels, improving the overall $\mathcal{J}\&\mathcal{F}$ by $21.6$ and $18.6$ points for the 4B and 8B models, respectively. These results indicate that the main improvement comes from Hi-GRPO's interactive clarification training rather than from the extra 120 samples alone.

\section{Sensitivity Analysis of $\lambda$.}\label{sec:sensitivity}
Table~\ref{tab:lambda} ablates the impact of the mixing coefficient $\lambda$, which balances the sequence-level holistic advantage and the step-level dense supervision. As shown in the table, $\lambda=0.5$ achieves the best performance. A smaller value ($\lambda=0.3$) allocates less weight to the step-level supervision, providing softer guidance that does not fully highlight the key reasoning steps in a long dialogue. Conversely, a larger value ($\lambda=0.7$) places more weight on the expert guidance. This tends to encourage the model to follow the teacher's output more closely, which may limit its exploration space and leads to sub-optimal final results. Therefore, $\lambda=0.5$ provides the optimal balance between the global outcome and the fine-grained refinement.
\begin{table}[h]
  \centering
  \begin{minipage}[t]{0.5\textwidth}
    \centering
    \vspace{-1em}
    \caption{Impact of the initial value of $\lambda$ in the Step-level Reward.}
    \label{tab:lambda}
    \setlength{\tabcolsep}{15pt}
    \footnotesize
    \begin{tabular}{c|ccc}
      \toprule
       $\lambda$ & $\mathcal{J}\&\mathcal{F}$ & $\mathcal{J}$ & $\mathcal{F}$ \\
      \midrule
      0.3 & 50.0 & 46.7 & 53.3 \\
      \rowcolor{rowhighlight}
      0.5 & \textbf{52.0} & \textbf{49.2} & \textbf{54.9} \\
      0.7 & 49.4 & 46.4 & 52.5 \\
      \bottomrule
    \end{tabular}
  \end{minipage}%
\end{table}

\section{Details of Efficiency Analysis}
\label{sec:efficiency_analysis}
Table~\ref{tab:latency_details} details inference efficiency and training costs. The ``Baseline'' denotes Qwen3-VL-4B in a non-interactive setting, where the model directly identifies and segments the target by itself. This make inference faster, but it lacks a mechanism to resolve the user's true intent and therefore obtains the worst performance. Among interactive variants, removing the turn-level or token-level optimization significantly increases the number of interaction turns at inference, while the segmentation performance steadily drops. This inverse trend proves our multi-level supervision prevents redundant questioning loops, enabling the model to ask highly discriminative questions and prune candidates efficiently. 
Regarding training, IC-Seg's overhead is highly manageable. Compared to the base trajectory-reward RL (221s per step), additional costs stem from invoking the expert guidance and an extra forward pass to compute token log-probabilities. Notably, the extra forward pass (IC-Seg vs. w/o step level) bypasses time-consuming text generation, increasing step time by merely $\sim 13\%$ (295s to 333.5s). Thus, our framework achieves substantial reasoning gains with minimal computational trade-offs.

\begin{table*}[h]
\setlength\tabcolsep{5pt}
\footnotesize
\centering
\caption{Detailed analysis of IC-Seg on Ambi-RVOS benchmark. ``Turns'' denotes the average per-sample query turns during inference. ``Time'' refers to the processing time(seconds) per sample. And ``Training cost'' indicates the average per-step time(seconds) during training.}
\label{tab:latency_details}
\begin{tabular}{l | ccc| ccc| ccc| c}
\toprule
\multirow{2}{*}{Method} & \multicolumn{3}{c|}{\textbf{Simple}} & \multicolumn{3}{c|}{\textbf{Medium}} & \multicolumn{3}{c|}{\textbf{Difficult}} & \multirow{2}{*}{\textbf{Training cost}} \\
\cmidrule(lr){2-10}
 & $\mathcal{J}\&\mathcal{F}$ & Turns & Time & $\mathcal{J}\&\mathcal{F}$ & Turns & Time & $\mathcal{J}\&\mathcal{F}$ & Turns & Time & \\
\midrule

\rowcolor{gray!7.5}
Baseline  & 35.7 & - & 1.2 & 24.6 & - & 0.9 & 17.3 & - & 0.8 & - \\
\midrule
\rowcolor{rowhighlight}
IC-Seg & 63.3 & 1.8 & 2.0 & 54.3 & 2.6 & 2.2 & 40.3 & 3.0 & 2.5 & 333.5 \\
w/o step level & 58.7 & 1.6 & 2.0 & 52.4 & 2.9 & 2.3 & 37.0 & 3.5 & 2.8 & 295.0 \\
w/o step \& turn level & 56.0 & 3.5 & 2.9 & 49.1 & 3.7 & 2.9 & 35.4 & 4.1 & 3.1 & 221.0 \\
\bottomrule
\end{tabular}
\end{table*}

\section{More Qualitative Analysis}

In Table \ref{tab:cot1}, we visualize the Chain-of-Thought (CoT) processes of IC-Seg in comparison with Qwen3-VL-4B* and Qwen3-VL-4B*+GRPO. Given the same initially ambiguous user query, Qwen3-VL-4B* tends to make arbitrary judgments, directly selecting the object it deems most suitable rather than proactively seeking clarification when faced with ambiguity. While Qwen3-VL-4B*+GRPO demonstrates the capability to ask clarifying questions, it frequently falls into repetitive questioning loops and ultimately fails after exceeding the maximum interaction limit. In contrast, IC-Seg exhibits superior performance in user interaction; it proactively queries the user to resolve ambiguities and avoids redundant questioning once the target is identified.

\begin{xltabular}{\textwidth}{X}
  \caption{Visualization of the CoTs between baselines and our IC-Seg. \label{tab:cot1}}\\
        \toprule
        \textbf{User Input:} Please segment "the person that spins and makes a slight jump."\\
        {\centering \includegraphics[width=\linewidth]{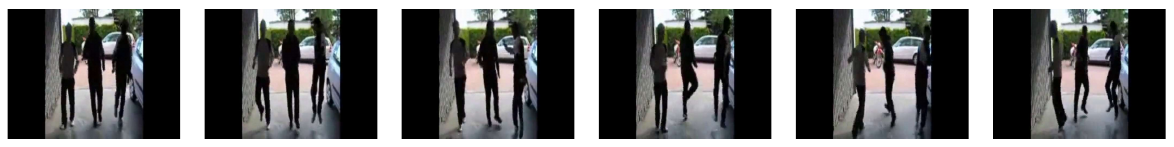}} \\
        \midrule
        \textbf{Qwen3-VL-4B*:}\\
        \cthink{<think>} There are three individuals in the frame, all appearing to be in motion. The central figure seems to be performing a more dynamic movement. \cthink{</think>}\\
        \ckey{<keyframe>} 210 \ckey{</keyframe>}\\
        Here is the high-res version of your selected frame:\\
        \includegraphics[width=3cm]{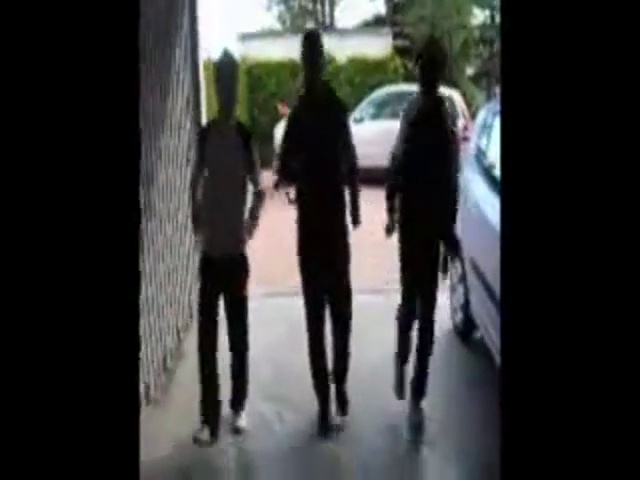} \\
        \cthink{<think>} The target is the person in the central, who is performing the spinning and jumping motion. The high-res frame clearly shows this individual, and the precise location should be centered on them for accurate identification. \cthink{</think>}\\
        \cans{<answer>} \{"bbox\_2d": [427, 57, 565, 906], "point\_2d": [496, 481]\} \cans{</answer>} \quad \rlap{\kern 5pt \smash{\raisebox{-0.5ex}{\textcolor{red}{\Huge \ding{55}}}}}
        \\
        \midrule

        \midrule
        \textbf{Qwen3-VL-4B*+\textit{GRPO}:}\\
        \cthink{<think>} There are three human figures in the video. All appear to be moving in a synchronized manner, possibly dancing or performing a routine. The central figure seems to be the most active, with a noticeable jump and spinning motion. The figure on the left is more static, while the one on the right is also moving but less dynamically than the center figure. To isolate the target, ask about the position relative to the others. \cthink{</think>}\\
        \csearch{<call>} \{"name": "vlm\_tool", "query": "Is the target the person in the middle, the one on the left, or the one on the right?"\} \csearch{</call>}\\
        \cinfo{<information>} the one on the right \cinfo{</information>}\\
        \cthink{<think>} Now that we know the target is the one on the right, we must verify if this figure is indeed performing the spinning and jumping motion. To confirm, we can ask about the specific action being performed by this figure. \cthink{</think>}\\
        \csearch{<call>} \{"name": "vlm\_tool", "query": "Is the target spinning and jumping, or just walking?"\} \csearch{</call>}\\
        \cinfo{<information>} spinning and jumping \cinfo{</information>}\\
        \cthink{<think>} The central figure is the only one who appears to be performing a more active motion — possibly the spinning and jumping. To resolve ambiguity, we must determine which person is performing the spinning and jumping motion. \cthink{</think>}\\
        \csearch{<call>} \{"name": "vlm\_tool", "query": "Is the target the person on the left, the one in the middle, or the one on the right?"\} \csearch{</call>}\\
        \cinfo{<information>} the one on the right \cinfo{</information>}\\
        \cthink{<think>} The target is confirmed to be the person on the right. However, the central figure is more active, but the query specifies the right person. To resolve this, we must focus on the specific motion — is the target performing a spin or jump? \cthink{</think>}\\
        \csearch{<call>} \{"name": "vlm\_tool", "query": "What is the target doing — is the person on the right spinning, jumping, or just walking?"\} \csearch{</call>}\\
        \cinfo{<information>} spinning \cinfo{</information>}\\
        \cthink{<think>} The target is still being described as the person on the right, but the central figure is the only one who appears to be performing a more dynamic motion. To resolve ambiguity, we must determine which person is performing the spinning and jumping motion. \cthink{</think>}\\
        \csearch{<call>} \{"name": "vlm\_tool", "query": "Is the target the person on the left, the one in the middle, or the one on the right?"\} \csearch{</call>}\\
        ... ...\\
        \texttt{MAX TURNS REACHED.}
        \quad \rlap{\kern 5pt \smash{\raisebox{-0.5ex}{\textcolor{red}{\Huge \ding{55}}}}}
        \\
        \midrule
        
        \textbf{IC-Seg-4B (ours): }\\
        \cthink{<think>} The target is the object that spins and makes a slight jump. The three individuals are visible, and the middle one is the most active, performing the spinning and jumping motion. The left and right individuals are relatively static. To isolate the target, I will ask whether the target is the person in the middle, left, or right. This will ensure the target is correctly identified without guessing. \cthink{</think>}\\
        \csearch{<call>} \{"name": "vlm\_tool", "query": "Is the target the person in the middle, the one on the left, or the one on the right?"\} \csearch{</call>}\\
        \cinfo{<information>} the one on the right \cinfo{</information>}\\
        \cthink{<think>} The target is confirmed to be the person on the right. The best frame to capture this target clearly is frame 295, as it shows the person on the right performing the spinning and jumping motion, with the motion being most visible.  \cthink{</think>}\\
        \ckey{<keyframe>} 295 \ckey{</keyframe>}\\
        Here is the high-res version of your selected frame:\\
        \includegraphics[width=3cm]{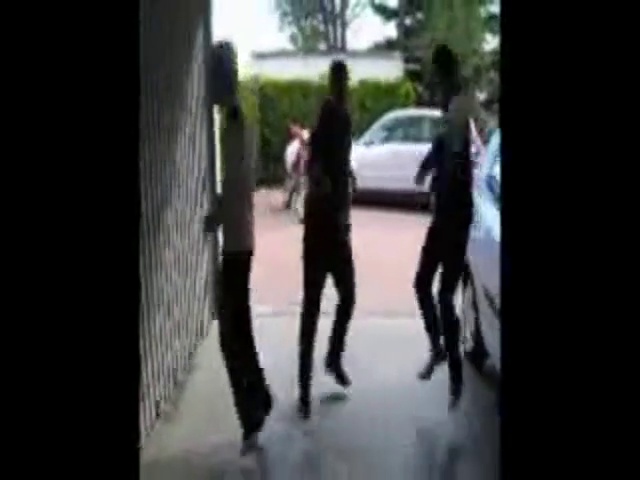} \\
        \cthink{<think>} The target is the person on the right, who is performing the spinning and jumping motion. The high-res frame clearly shows this individual, and the precise location should be centered on them for accurate identification. The bounding box should encompass the person on the right, and the point should be placed on them to confirm the target. \cthink{</think>}\\
        \cans{<answer>} \{"bbox\_2d": [643, 88, 758, 853], "point\_2d": [700, 500]\} \cans{</answer>} \quad \rlap{\kern 5pt \smash{\raisebox{-0.5ex}{\textcolor{green!60!black}{\Huge \ding{52}}}}}
        \\
        \hline
\end{xltabular}

\section{Prompts Details} \label{sec:prompt}

\begin{tcolorbox}[
    colback=white,
    colframe=gray!85,
    coltitle=white,
    fonttitle=\bfseries,
    title={System Prompt for IC-Seg},
    boxrule=0.8pt,
    arc=3pt,
    left=5pt, right=5pt, top=5pt, bottom=5pt,
    breakable
]
Your role is a video target identification assistant. 
You will be given an object query and a sequence of video frames. Each frame is preceded by its index. 
Your task is to locate the target with a bounding box and a point in your selected frame.

\vspace{1em}
\# Tools

You have access to the following tool:

\vspace{1em}
\csearch{<tools>}

\{``name'': ``vlm\_tool'', ``description'': ``Ask for clarification when multiple candidates match the existing information.'', ``parameters'': \{``query'': \{``type'': ``string'', ``description'': ``The clarification question.''\}\}

\csearch{</tools>}

The tool will return the answer between \cinfo{<information>} and \cinfo{</information>}.

\vspace{1em}
\# Output Format:
Depending on the situation, output one of the following:

\vspace{.5em}
\#\# 0. When the target is ambiguous (multiple candidates):

\cthink{<thinking>} 

(1) Analyze Candidates: Identify all candidates that currently match the existing description; (2) Feature Extraction: Identify key visual features (color, pose, position, etc.) that differ among these candidates; (3) Formulate Question: Select the question that most evenly splits or most effectively isolates the candidates, and ask about that specific attribute (e.g., color, direction, action, relative position, shape). 

\cthink{</thinking>}

\csearch{<call>} \{``name'': ``vlm\_tool'', ``query'': <string>\} \csearch{</call>}

\vspace{.5em}
\#\# 1. When the target is uniquely identified:

\cthink{<thinking>}

Analyze the provided frames and select the best one where the target is most clearly visible

\cthink{</thinking>}

\ckey{<keyframe>} [integer frame\_index] \ckey{</keyframe>}

\vspace{.5em}
\#\# 2. Once you receive the high-res keyframe:

\cthink{<thinking>} 

(1) Describe the unique visual features of the target to prove you captured it; (2) Determine the precise 2D location of the target with bbox and the point inside the object. 

\cthink{</thinking>}

\cans{<answer>} \{``bbox\_2d'': [x1,y1,x2,y2], ``point\_2d'': [x,y]\} \cans{</answer>}

\vspace{1em}
\# IMPORTANT NOTE

At each step, 

1. You are STRICTLY FORBIDDEN from guessing or picking an option when ambiguity exists.

2. Always look at the video frames first before you believe the target is uniquely identified. If after watching the video, you find that the target is not unique and you need to call the 'vlm\_tool'.

3. Do not put the original query inside your question. Use ``the target'' instead.

4. Ask about exactly one visual attribute (color / direction / action / relative position / shape). For static questions, especially absolute position,consider including a specific frame number when asking the question so that vlm\_tool can answer the question more accurately.

5. You are FORBIDDEN from using 'vlm\_tool' to ask which frame is the clearest or most visible. You must make this judgment yourself based on the provided video frames.

\vspace{1em}
\# Note: Always use the viewer's perspective for left/right orientation

\end{tcolorbox}

\begin{tcolorbox}[
    colback=white,
    colframe=gray!85,
    coltitle=white,
    fonttitle=\bfseries,
    title={System Prompt for User Simulator in Answering Questions},
    boxrule=0.8pt,
    arc=3pt,
    left=5pt, right=5pt, top=5pt, bottom=5pt,
    breakable
]
You are an expert visual analysis assistant. Your task is to accurately answer a query regarding a specific target object across a sequence of Numbered Video Frames.

\vspace{1em}
\# Context \& Target:

- A sequence of indexed video frames.

- The specific object you must track and analyze is enclosed within a RED CONTOUR. The target will remain unchanged throughout the entire process. Do not assume that the target corresponding to the red outline has changed due to camera movement or obstruction.

- We also provide text descriptions of the object with red outline to help you locate the specific target and answer query.

- The query requires understanding the object's appearance, temporal dynamics, movements, or state changes across the provided frames.

\vspace{1em}
\# Strict Rules \& Logic:

1. You must strictly focus on the object inside the red contour. If there are other similar or identical objects in the image, you MUST IGNORE THEM. Your answer must apply ONLY to the contoured target, never mixing its attributes with others.

2. You must track the red contour across all provided frames. The query is designed to be answered by observing the entire temporal sequence.

3. You are ABSOLUTELY FORBIDDEN from revealing any additional attributes, colors, actions, or context about the target that were not explicitly asked for. Answer ONLY the specific question posed.

4. If a question is ambiguous or cannot be answered definitively, provide a clear indication and request clarification.

\vspace{1em}
\# Output Format:
You must strictly output your reasoning in \cthink{<thinking>} tags, followed by your final concise answer in \cans{<answer>} tags.

\vspace{.5em}
\cthink{<thinking>} 

1. Track the object enclosed in the RED CONTOUR from the first frame to the last.

2. Synthesize the object's action, movement, or interaction across the timeline.

3. Formulate the absolute minimal text needed to answer the query. If none fit the observed events, conclude ``neither''.

\cthink{</thinking>}

\vspace{.5em}
\cans{<answer>} 

Provide a concise answer (e.g., ``it moved to the table'', ``the red one'', ``yes'', ``no'', ``neither''...). 

- Do NOT mention ``red contour'', ``red box'', or provide unnecessary explanations in this tag.

\cans{</answer>}

\vspace{1em}
\# Note: Always use the viewer's perspective for left/right orientation
\end{tcolorbox}

\begin{tcolorbox}[
    colback=white,
    colframe=gray!85,
    coltitle=white,
    fonttitle=\bfseries,
    title={System Prompt for User Simulator in Guidance Generation},
    boxrule=0.8pt,
    arc=3pt,
    left=5pt, right=5pt, top=5pt, bottom=5pt,
    breakable
]
You are an expert Video Object Reasoning and Filtering Tutor. Your goal is to analyze interactive queries and environment answers to provide generalized, a priori strategic guidance focused on efficient search space reduction.

\vspace{1em}
\# Inputs:

(1). A sequence of indexed video frames.

(2). The initial user query and total matching candidate objects (initial\_count).

(3). Descriptions of all initial qualified objects.

(4). Dialogue Sequence: A JSON-formatted list of tool-use interactions.

\vspace{1em}
\# Task:

1. Calculate Target Subsets: Determine the remaining candidate count after each dialogue turn.

2. Formulate Holistic Guidance: Abstract the reasoning trajectory into an objective, forward-looking tactical manual.

   \vspace{.5em}
   - Extract Principles: Translate successful filtering actions into general declarative rules about candidate space reduction.

   \vspace{.5em}
   - Correct Inefficiencies: Translate redundant steps into proactive optimization rules for better information gain (e.g., "Query multi-axis positions simultaneously to eliminate larger subsets, rather than verifying single axes").

   \vspace{.5em}
   - Address Unresolved Ambiguity: If the dialogue ends with multiple candidates remaining, specify the discriminative dimensions (e.g., relative spatial relations, ordinal ranking among the remaining options, nuanced dynamic interactions) required to further differentiate the remaining pool.

   \vspace{.5em}
   - CRITICAL CONSTRAINTS:
   
     - Absolutely forbid referencing the current interaction. Do NOT use phrases like "In this case", "Your query", "The student did", or "As seen previously", etc.
     
     - Do NOT claim a strategy will "isolate the target" or "find the correct object", as the ground truth is unknown. Focus on "disambiguation" and "narrowing candidates".
     
     - Every sentence MUST be a concrete, executable tactic. Forbid empty praise or concluding remarks.
     
     - Do NOT quote specific queries, exact responses, or mention the "red outline".

\vspace{1em}
\# Output Format:

\vspace{.5em}
\cthink{<thinking>}

1. Identify initial candidate objects based on inputs.

2. Sequential Filtering: Iterate through the Dialogue Sequence to identify remaining objects per turn.

3. Analyze trajectory to form holistic advice, strictly adhering to constraints.

\cthink{</thinking>}

\vspace{.5em}
\cans{<output>}

\{
  
  ``sequential\_subset\_count'': <list of int>, \/\/ Starts with initial\_count, followed by count after each turn. Non-increasing. Length = len(dialogue) + 1.
  
  ``holistic\_guidance'': <string> \/\/ A comprehensive, objective paragraph of tactical principles focusing on disambiguation and space pruning, completely stripped of past references and filler.
  
\}

\cans{</output>}
\end{tcolorbox}

\section{User Study Details}\label{sec:user}
Our user study involved five participants. They were instructed to: (1) observe a video containing an ambiguous query, where the ground-truth (GT) target was specifically highlighted with a red outline for clarity; and (2) provide factual responses to the model's clarifying questions based on the actual status and attributes of the marked GT target. All participants were volunteers from the local academic community and were fully informed about the research's purpose.

\section{Limitations and Future Work} \label{sec:limitation}
Despite the effectiveness of IC-Seg, one current limitation is that the system primarily supports text-based interaction. While text is highly expressive for clarification, in many real-world scenarios, human intent can be conveyed more intuitively through other modalities. Future research could explore expanding the framework to support multi-modal interaction, such as incorporating pointing gestures or audio-visual cues to resolve ambiguities more efficiently. Integrating these diverse interaction signals would move language-guided segmentation toward a more natural and seamless human-AI collaborative experience.

\section{Broader Impact} \label{sec:impact}
In this paper, we studied intent clarification for language-guided segmentation, introducing IC-Seg—an agentic framework that proactively resolves ambiguity through interaction. Beyond its technical contributions, this work has broader societal implications for human-AI collaboration. By enabling models to "ask" rather than "guess," IC-Seg can significantly enhance the safety and reliability of interactive AI systems in critical domains, such as assistive robotics and autonomous navigation, where misinterpreting human intent could lead to physical or operational risks.

Furthermore, this interactive paradigm promotes the democratization of AI by lowering the barrier for non-expert users to collaborate with complex visual systems through intuitive, multi-turn dialogue. Therefore, our approach not only mitigates the risks of erroneous task execution but also significantly reduces the cognitive burden on users by alleviating the pressure to provide precise initial prompts.


\end{document}